%% file: main.tex
\documentclass[11pt,a4paper,abstract=on]{scrartcl}

\usepackage[utf8]{inputenc}
\usepackage[english]{babel}

\usepackage[paper=a4paper,left=25mm,right=25mm,top=25mm,bottom=30mm]{geometry}
\usepackage{graphicx}
\usepackage{latexsym}
\usepackage{amsmath,amssymb,amsthm,dsfont}
\usepackage{bbm}
\usepackage{wrapfig}
\usepackage{scrlayer-scrpage}
\usepackage{adjustbox}
\usepackage{tabularx}
\usepackage{mathtools}
\usepackage{booktabs}
\usepackage{authblk}
\usepackage{capt-of}
\usepackage{natbib}
\usepackage[breaklinks, colorlinks=true, citecolor=black, urlcolor=black, linkcolor=black]{hyperref}
\usepackage[table]{xcolor}

\usepackage{natbib}
\usepackage{etoolbox}

\makeatletter
\newcommand\bibstyle@comma{\bibpunct(),a,,}
\newcommand\bibstyle@semicolon{\bibpunct();a,,}
\makeatother

\pretocmd\citet{\citestyle{comma}}\relax\relax
\pretocmd\citep{\citestyle{semicolon}}\relax\relax

\numberwithin{equation}{section} 

\newcommand{\R}{\mathbb{R}} % real
\newcommand{\N}{\mathbb{N}} % natural

\newcommand{\Vaw}{V$_a^{w}$}
\newcommand{\Va}{V$_a^{=}$}
\newcommand{\Vw}{V$_0^{w}$}
\newcommand{\Vo}{V$_0^{=}$}

\begin{document}

\title{Aggregating distribution forecasts from \\ deep ensembles}

\author[1]{Benedikt Schulz}
\author[1]{Lutz K\"ohler}
\author[1,2]{Sebastian Lerch}
\affil[1]{Karlsruhe Institute of Technology}
\affil[2]{Heidelberg Institute for Theoretical Studies}
\renewcommand\Authands{ and }
	
\date{\today}

\maketitle

\begin{abstract}
\noindent 
The importance of accurately quantifying forecast uncertainty has motivated much recent research on probabilistic forecasting. In particular, a variety of deep learning approaches has been proposed, with forecast distributions obtained as output of neural networks. These neural network-based methods are often used in the form of an ensemble, e.g., based on multiple model runs from different random initializations or more sophisticated ensembling strategies such as dropout, resulting in a collection of forecast distributions that need to be aggregated into a final probabilistic prediction. With the aim of consolidating findings from the machine learning literature on ensemble methods and the statistical literature on forecast combination, we address the question of how to aggregate distribution forecasts based on such `deep ensembles’. Using theoretical arguments and a comprehensive analysis on twelve benchmark data sets, we systematically compare probability- and quantile-based aggregation methods for three neural network-based approaches with different forecast distribution types as output. Our results show that combining forecast distributions from deep ensembles can substantially improve the predictive performance. We propose a general quantile aggregation framework for deep ensembles that
allows for corrections of systematic deficiencies and performs well in a variety of settings, often superior compared to a linear combination of the forecast densities. Finally, we investigate the effects of the ensemble size and derive recommendations of aggregating distribution forecasts from deep ensembles in practice.
\end{abstract}

\section{Introduction} \label{sec:introduction}

Probabilistic forecasts in the form of predictive probability distributions over future quantities or events aim to quantify the uncertainty in the predictions and are essential to optimal decision making in applications \citep{Gneiting2014,KendallGal2017}.
Motivated by their superior performance on a wide variety of machine learning tasks, much recent research interest has focused on the use of deep neural networks (NNs) for probabilistic forecasting. 
Different approaches for obtaining a forecast distribution as the output of a NN have been proposed over the past years, including parametric methods where the NN outputs parameters of a parametric probability distribution \citep{Lakshminarayanan2017,DInsantoPolsterer2018,Rasp2018}, semi-parametric approximations of the quantile function of the forecast distribution \citep{Bremnes2020} and nonparametric methods where the forecast density is modeled as a histogram \citep{Gasthaus2019,Li2021}.
To account for the randomness of the training process based on stochastic gradient descent methods, NNs are often run several times from different random initializations. 
\citet{Lakshminarayanan2017} refer to this simple to implement and readily parallelizable approach as deep ensembles (DEs).
We will adopt the term deep ensemble to refer to ensembles of NN predictions in general, independent of the ensemble generating mechanism.\footnote{Note that our terminology thus differs from \citet{Lakshminarayanan2017}, who introduced the term DE exclusively for ensembles of NNs generated based on random initialization.}
Other than random initialization, more sophisticated approaches for the generation of DEs have been proposed with dropout \citep{Srivastava2014} being a prominent example.
DEs for probabilistic forecasting thus yield an ensemble of predictive probability distributions.
To provide a final probabilistic forecast, the ensemble of predictive distributions needs to be aggregated to obtain a single forecast distribution.

The task of combining predictive distributions has been studied extensively in the statistical literature, see \citet{Gneiting2013}, \citet[Section 2.6]{PetropoulosEtAl2022} and \citet{Wang2023} for overviews.
Combining probabilistic forecasts from different sources has been successfully used in a wide variety of applications including economics \citep{AastveitEtAl2019}, epidemiology \citep{Cramer2022,TaylorTaylor2021}, finance \citep{Berkowitz2001}, signal processing \citep{KolianderEtAl2022} and weather forecasting \citep{BaranLerch2016,BaranLerch2018}, and constitutes one of the typical components of winning submissions to forecasting competitions \citep{BojerMeldgaard2021,JanuschowskiEtAl2021}. 
On the other hand, forecast combination also forms the theoretical framework of some of the most prominent techniques in machine learning such as boosting \citep{Freund1996}, bagging \citep{Breiman1996} or random forests \citep{Breiman2001}, which are based on the idea of building ensembles of learners and combining the associated predictions. 
Generally, the individual component models (or ensemble members) can be based on entirely distinct modeling approaches, or on a common modeling framework where the model training is subject to different input data sets of other sources of stochasticity. 
The latter is the case for the DEs we consider in this study.
For general reviews on ensemble methods in machine learning, we refer to \citet{Dietterich2000}, \citet{ZhouEtAl2002} and \cite{Ren2016}.

While the arithmetic mean is a powerful and widely accepted method for aggregating single-valued point forecasts, the question how probabilistic forecasts should be combined is more involved and has been a focus of research interest in the literature on statistical forecasting \citep{Gneiting2013,PetropoulosEtAl2022}.
We will focus on readily applicable aggregation methods for the combination of probabilistic forecasts from DEs.
A widely used approach is the linear aggregation of the forecast distributions, which is often referred to as linear (opinion) pool (LP). 
An alternative is given by aggregating the forecast distributions on the scale of quantiles by linearly combining the corresponding quantile functions, an approach that is commonly referred to as Vincentization \citep[VI; see, for example,][]{Genest1992} dating back to \citet{Vincent1912}.
In particular, the LP and VI have been compared to each other coming from a theoretical perspective in \citet{Lichtendahl2013} and \citet{Busetti2017}, and from a practical perspective in an application of DEs for electricity price forecasting in \citet{Marcjasz2023}.

In recent years, there has been a surge in interest in DEs in general, see \citet{Ganaie2022} and \citet{Mohammed2023} for overviews. 
In particular, DEs of distributional forecasts have been investigated in various studies, e.g., 
\citet{Lakshminarayanan2017} propose the LP in combination with parametric densities from a DE as alternative to Bayesian NNs, and
\citet{Rahaman2020} investigate the uncertainty quantification properties of probabilistic DEs.
Further, \citet{Fakoor2023} develop a VI framework for aggregating quantile regression models, e.g., based on DEs, using methods from deep learning. 
In contrast to their work, we focus only on DEs and full distributional forecasts, which can be  defined arbitrarily.

This study is motivated by and based on our work in \cite{Schulz2022}, where we use DEs to statistically postprocess probabilistic forecasts for the speed of wind gusts and propose a common framework of NN-based probabilistic forecasting methods with different types of forecast distributions.
In the following, we apply a two-step procedure by first generating an ensemble of probabilistic forecasts and then aggregating them into a single final forecast, which matches the typical workflow of forecast combination from a forecasting perspective. 
Alternatively, it is also possible to incorporate the aggregation procedure directly into the model estimation \citep{Fakoor2023}.

\subsection{Contribution} \label{ssec:contribution}

The main aim of our work is to consolidate findings from the statistical and machine learning literature on forecast combination and ensembling for probabilistic forecasting.
Our study is the first to systematically investigate and compare the two central aggregation schemes for probabilistic forecasts, namely, probability (LP) and quantile aggregation (VI), applied to DEs.
In addition to the LP and the standard approach to VI, we propose a novel general VI approach that is able to correct for systematic errors such as biases and miscalibration in the aggregated forecasts.
Using theoretical arguments and a comprehensive evaluation on machine learning benchmark data sets, we analyze the aggregation methods with different ways to characterize the corresponding forecast distributions and different ensembling strategies.
Our findings include advice on the choice of the most suitable aggregation method based on theoretical arguments, tailored to chosen type of distribution forecasts, and the application for different NN methods, ensembling strategies and data sets of varying complexity.

\subsection{Outline} \label{ssec:outline}

The remainder of the paper is organized as follows. 
Section \ref{sec:aggregation} introduces relevant metrics for evaluating probabilistic forecasts and the forecast aggregation methods.
Three NN-based methods for probabilistic forecasting are presented in Section \ref{sec:prob_networks} along with a discussion of how the different aggregation methods can be used to combine the corresponding predictive distributions of an ensemble of such forecasts. 
Section \ref{sec:prob_networks} ends with a short introduction of the strategies used for the generation of DEs.
In Section \ref{sec:cs}, we apply the aggregation methods in a comprehensive case study. First, we provide an in-depth analysis for two selected pairs of data set and ensembling strategy, then we compare the performance for all data sets and ensembling strategies.
Section \ref{sec:conclusion} concludes with a discussion.
Code with implementations of all methods is available online (\url{https://github.com/benediktschulz/ADDE}).

\section{Combining probabilistic forecasts} \label{sec:aggregation}

Probabilistic forecasts given in the form of predictive probability distributions for future quantities or events aim to quantify the uncertainty inherent to the prediction. In the following, we first summarize how such distribution forecasts can be evaluated, and then formally introduce the LP and VI methods for aggregating probabilistic forecasts.

\subsection{Assessing predictive performance} \label{ssec:evaluation}

In our evaluation of predictive performance, we will follow the principle of \cite{Gneiting2007probabilistic} that a probabilistic forecast should aim to maximize sharpness subject to calibration. Calibration refers to the statistical consistency between the forecast distribution and the observation, whereas sharpness is a property of the forecast alone and refers to the degree of forecast uncertainty. A forecast is said to be sharper, the smaller the associated uncertainty.

Quantitatively, calibration and sharpness can be assessed simultaneously using proper scoring rules \citep{Gneiting2007scoring}. A scoring rule $S(F,y)$ assigns a penalty to a pair of a probabilistic forecast $F$ and corresponding observation $y \in \R$ and is called proper if the underlying true distribution $G$ scores lowest in expectation.
Our forecast evaluation in the following will mainly focus on the widely used continuous ranked probability score \citep[CRPS;][]{Matheson1976} 
\begin{equation}
    	\text{CRPS} (F, y) = \int_{-\infty}^{\infty} \left( F(z) - \mathbbm{1} \lbrace y \leq z \rbrace \right)^2 dz, \quad y \in \R,
 \label{eq:crps}
\end{equation}
where $F$ is a forecast distribution with finite first moment and $\mathbbm{1}$ is the indicator function. Proper scoring rules such as the CRPS are not only used for forecast evaluation but also provide valuable tools for estimating model parameters. In the case of the CRPS, the estimation typically relies on closed-form analytical expressions of the integral in \eqref{eq:crps} \citep[see, for example,][]{Jordan2019scoringrules} and is referred to as optimum score estimation \citep{Gneiting2007scoring}.

To compare competing forecasting methods based on a proper scoring rule with respect to a reference, we calculate the associated skill score, here the continuous ranked probability skill score (CRPSS), 
which is defined as the relative improvement over the reference method. 
Let $\bar S_{\text{f}}$ denote the mean score of the forecasting method of interest over a given data set and $\bar S_{\text{ref}}$ the corresponding mean score of the reference forecast. The associated skill score $SS_{\text{f}}$ is then calculated via
\begin{eqnarray}
SS_{\text{f}} = 1 - \dfrac{ \bar S_{\text{f}} }{ \bar S_{\text{ref}}}. \label{eq:crpss}
\end{eqnarray}
In contrast to proper scoring rules, skill scores are positively oriented with 1 indicating optimal predictive performance, 0 no improvement over the reference and a negative skill a decrease in performance. 

Further, we assess calibration qualitatively via histograms of the probability integral transform (PIT), 
% $F(Y)$. 
which is defined as the value of the CDF at the observation.\footnote{Technically, we here use the unified PIT, a generalization proposed in \citet{Vogel2018}, due to the format of some of the aggregated forecast distributions.}
A probabilistic forecast is (well-)calibrated, if the PIT is uniformly distributed, resulting in a flat histogram. An U-shaped PIT histogram indicates underdispersion (or overconfidence), that is, a lack of spread in the forecast distribution, whereas a hump-shaped histogram indicates overdispersion (or underconfidence), that is, too much spread. 
In mathematical terms, the dispersion of a predictive distribution can be defined as the variance of the PIT. For a calibrated forecast, the variance should equal that of an uniform distribution, i.e., 1/12 $\approx$ 0.0833. A variance smaller than 1/12 corresponds to overdispersion, a variance larger than 1/12 to underdispersion.
In addition, we generate quantile-based prediction intervals (PIs) to assess the calibration of the forecast distributions via the empirical coverage, and the sharpness via the length of the PIs. If a forecast is well-calibrated, the empirical coverage should resemble the nominal coverage, and a forecast is the sharper, the smaller the length of the PI. 
The nominal level of the PIs is a tuning parameter for evaluation, which we choose to be 90\% for the case study in Section \ref{sec:cs}. 
For further background and details on the assessment of probabilistic forecasts, we refer to \citet[][Appendix A]{Schulz2022} and the references therein.

At last, we briefly address how we measure diversity within an ensemble of predictive distributions. We differentiate between three different measures, namely, in terms of the location, prediction uncertainty and performance. For the location, we calculate the mean of each predictive distribution and then use the standard deviation of these mean values from all ensemble members as measure of the location diversity. We proceed analogously for the prediction uncertainty by using the PI length instead of the mean, or the CRPS for the performance respectively. As the standard deviation is scale-dependent and the goal is to compare among different settings, we standardize the diversity values within one set of predictions and calculate the mean to obtain one summarizing value.

\subsection{Combining predictive distributions} \label{ssec:agg_methods}

Given $n \in \N$ individual probabilistic forecasts we aim to aggregate, we will denote their cumulative distribution functions (CDFs) by $F_1, \dots, F_n$ and their quantile functions by $Q_1, \dots, Q_n$.
In the following, 
the aggregation methods introduced below will typically assign weights $w_1$, \dots, $w_n$ to the individual forecast distributions. 
We apply the aggregation methods to forecasts produced by the same data-generating mechanism based on a DE. Therefore, we do not expect systematic differences between the individual forecasts and only consider equally weighted ensemble members in the following.

\subsubsection{Linear pool (LP)} \label{sssec:agg_linear_pool}

The most widely used approach for forecast combination is the LP, which is the arithmetic mean of the individual forecasts \citep{Stone1961}. For probabilistic forecasts, the LP is calculated as the (in our case equally) weighted average of the predictive CDFs and results in a mixture distribution. Equivalently, the LP can be calculated by averaging the probability density functions (PDFs). We define the predictive CDF of the LP via
\begin{eqnarray}
	F_w (z) := \sum_{i = 1}^{n} w_i F_i (z), \quad z \in \R, \label{eq:linear_pool}
\end{eqnarray}
where $w_i \geq 0$ for $i = 1, \dots, n$ with $\sum_{i = 1}^{n} w_i = 1$. Note that the weights need to sum up to 1 to ensure that $F_w$ yields a valid CDF.
Hence, our assumption of equal weights results in the choice of $w_i = \frac{1}{n}$ for $i = 1, \dots, n$ in \eqref{eq:linear_pool}.

The LP has some appealing theoretical properties\footnote{For example, \citet{Lichtendahl2013} and \citet{Abe2022} show that the score of the LP forecast is at least as good as the average score of the individual components in terms of different proper scoring rules.} and has been the prevalent forecast aggregation method over the last decades. For example, \cite{Lakshminarayanan2017} use the LP to combine density forecasts of multiple NNs.
However, there are disadvantages to the use of the LP that is known to have suboptimal properties when aggregating probabilities, since a linear combination of probability forecasts results in less sharp and more underconfident forecasts \citep{Ranjan2010}. \cite{Gneiting2013} extend this result to the general case of predictive distributions by showing that in case of distribution forecasts sharpness decreases and dispersion increases. In particular, a (non-trivial) combination of calibrated forecasts is not calibrated anymore.
In the context of DEs, these downsides have also been observed in recent studies \citep{Rahaman2020,Wu2021}.

\begin{figure}
\begin{center}
	\includegraphics[width=\textwidth]{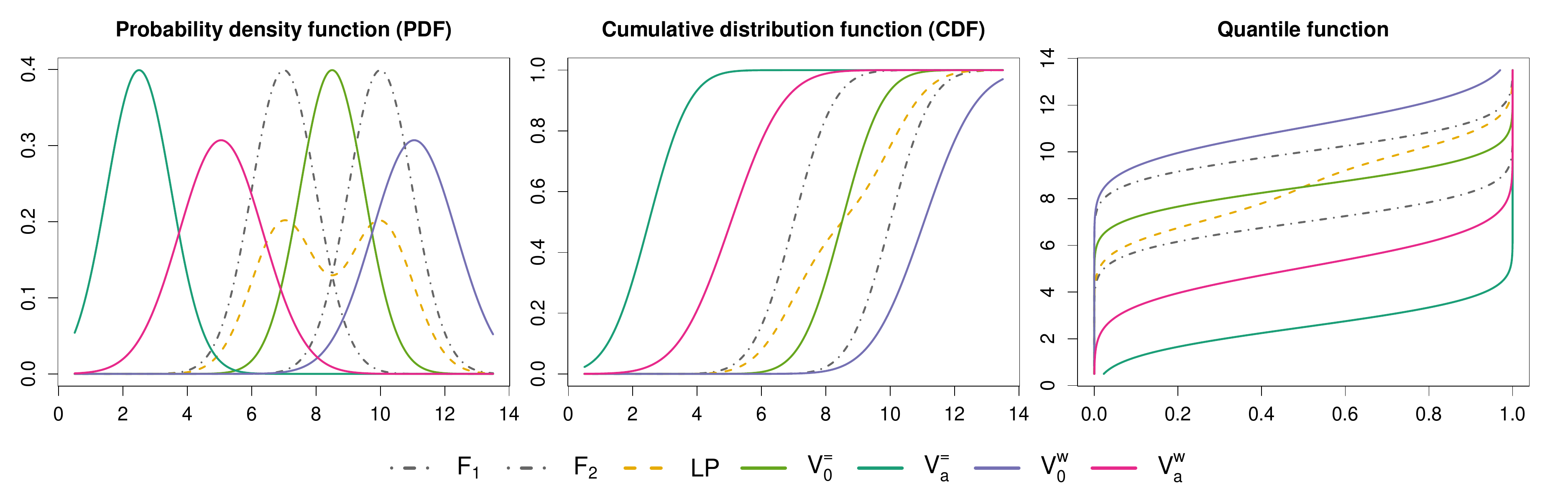}
	\caption{PDF, CDF and quantile function of two normally distributed forecasts $F_1$ and $F_2$ ($\mu_1 = 7$, $\mu_2 = 10$, $\sigma_1 = \sigma_2 = 1$) together with forecasts aggregated via the methods presented in Section \ref{sec:aggregation}. \Va\ and \Vaw\ use the intercept $a = -6$, \Vw\ and \Vaw\ the weight $w_0 = 0.65$. \label{fig:agg_methods}}
\end{center}
\end{figure}

Figure \ref{fig:agg_methods} illustrates the effect of forecast combination via the LP for two exemplary normal distributions.
The aggregated forecasts is a bimodal distribution, which is less confident, i.e., more spread out, than the individual members.
In case of overconfident forecasts, which are often generated by NN models, this increase in spread typically improves predictive performance. However, in case of calibration or underconfidence, the forecasts become less well calibrated.

\subsubsection{Vincentization (VI)} \label{sssec:agg_vincentization}

While the LP aggregates the forecasts on a probability scale, VI performs a quantile-based linear aggregation \citep{Ratcliff1979,Genest1992}. We extend the standard VI framework\footnote{To the best of our knowledge, VI is usually only applied 
with non-negative weights that sum up to 1 and without an intercept 
\citep[e.g.,][]{Fakoor2023}.
Exceptions include \citet{Wolffram2021} and related, unpublished simulation experiments by Anja Mühlemann (University of Bern, 2020, personal communication). 
} by defining the VI quantile function via
\begin{eqnarray}
	Q^a_w (p) := a + \sum_{i = 1}^{n} w_i Q_i (p), \quad p \in \left[0, 1\right], \label{eq:vincentization}
\end{eqnarray}
where $a \in \R$ and $w_i \geq 0$ for $i = 1, \dots, n$. In contrast to the LP, the weights do not need to sum to 1 and only their non-negativity is required to ensure the monotonicity of the resulting quantile function $Q^a_w$. 
Again, we will consider only the case of equal weights, which here translates to a free weight parameter $w_i = w_0 \geq 0$ for $i = 1,...,n$.
Further, a real-valued intercept $a$ is added to the aggregated quantile functions to correct for systematic biases.

Given equal weights, we consider four different variants of VI. First, with weights that sum up to 1 and no intercept, that is, $a = 0$ and $w_0 = \frac{1}{n}$, which is referred to by \Vo. Similar to the LP,  \Vo\ does not require the estimation of any parameters.
Further, we consider VI variants where we estimate the parameters $a$ and $w_0$ both independently (while the other is fixed at the values of \Vo) and also simultaneously, resulting in the three variants \Va\ (where $w_0 = \frac{1}{n}$ and $a$ is estimated), \Vw\ (where $a = 0$ and $w_0$ is estimated) and \Vaw\ (where both $a$ and $w_0$ are estimated). The parameters are estimated minimizing the CRPS following the optimum scoring principle. The standard procedure for training machine learning models where the available data is split into training, validation and test data sets offers a natural choice for estimating the combination parameters. Given NN models estimated based on the training set (where the validation set is used to determine hyperparameters), we estimate the coefficients of the VI approaches separately in a second step based on the validation set, which can be seen as a post-hoc calibration step \citep{Guo2017}. During this second step, the component models with quantile functions $Q_i, i = 1,\dots,n,$ are considered fixed and we only vary the combination parameters in \eqref{eq:vincentization}. In the following, we will restrict our attention to fixed training and validation sets, but an extension of the approach described here to a cross-validation setting is straightforward. Table \ref{tbl:agg_methods} provides an overview of the abbreviations and important characteristics of the different forecast aggregation methods we will consider below.

\begin{table}
\caption{Overview of the aggregation methods for probabilistic forecasts, with $F_i$ and $Q_i$ denoting the predictive CDFs and quantile functions of the individual components models. The column `Parameters' indicates which parameters are estimated based on data, following the procedure described in Section \ref{sssec:agg_vincentization}. \label{tbl:agg_methods}}	
\begin{center}
\begin{tabular}{l@{\hskip 0.5cm}l@{\hskip 0.5cm}l@{\hskip 0.75cm}l@{\hskip 0.75cm}l}
	\toprule 
	Abbr.\ & Scale & Formula & Parameters & Estimation  \\
	\midrule
	LP & Probability & $F_w = \frac{1}{n} \sum_{i = 1}^{n} F_i$ & - & - \\ 
	\midrule
	\Vo\ & Quantile & $Q_w = \frac{1}{n} \sum_{i = 1}^{n} Q_i$ & - & - \\
	\Va\ & Quantile & $Q_w = \frac{1}{n} \sum_{i = 1}^{n} Q_i + a$ & $a \in \R$ & CRPS \\
	\Vw\ & Quantile & $Q_w = w_0 \sum_{i = 1}^{n} Q_i$ & $w_0 \geq 0$ & CRPS \\
	\Vaw\ & Quantile & $Q_w = w_0 \sum_{i = 1}^{n} Q_i + a$ & $w_0 \geq 0$, $a \in \R$ & CRPS \\
	\bottomrule
% \end{tabularx}
\end{tabular}
\end{center}
\end{table}

While the effects of the LP on calibration and dispersion have been proven mathematically, no such strong statements for the VI exist.
\citet{Lichtendahl2013}, who compare the theoretical properties of the LP and \Vo, note that the aggregated predictive distributions both yield the same mean but the VI forecasts are sharper, that is, the VI predictive distribution has a variance smaller or equal to that of the LP.
Related work in the statistical literature includes comparisons to the LP which demonstrate that VI tends to perform better than the LP \citep{Lichtendahl2013,Busetti2017}. 

Figure \ref{fig:agg_methods} illustrates the effects of VI in the exemplary case of two normal distributions.
To highlight the influence of the individual VI parameters, we note that the intercept $a$ only has an effect on the location of the resulting aggregated distribution, e.g., the predictive density of \Va\ in Figure \ref{fig:agg_methods} is shifted along the x-axis. 
In contrast, the weight $w_0$ has an effect on both the location and the spread. If it is larger than $\frac{1}{n}$, the spread increases compared to the average spread of the individual forecasts 
(as in Figure \ref{fig:agg_methods} for \Vw\ and \Vaw), 
and it decreases for values smaller than $\frac{1}{n}$. 
However, a weight not equal to 1 also shifts the location of the distribution,
as we can see for \Vw\ in Figure \ref{fig:agg_methods}. 
At last, we can control both the location and the spread of the aggregated distribution by choosing a weight and an intercept using \Vaw.

\section{Distribution forecasts from deep ensembles} \label{sec:prob_networks}

In this section, we will specify how we generate distribution forecasts from DEs. First, we will introduce the three NN approaches used to generate distributional forecasts and how to apply the aggregation methods presented in Section \ref{ssec:agg_methods}. Then, we will present the different ensembling strategies we consider for the generation of DEs.

\subsection{Neural network methods for probabilistic forecasting} \label{ssec:prob_networks_distr}

\begin{figure}
\begin{center}
	\includegraphics[width=0.82\textwidth]{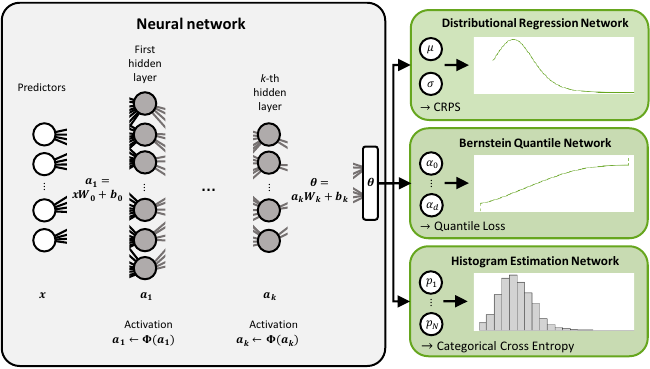}
	\caption{Graphical illustration of the general framework for NN-based probabilistic forecasting. \label{fig:nn_diagram}}
\end{center}
\end{figure}

In the context of probabilistic wind gust prediction, \citet{Schulz2022} propose a framework for NN-based probabilistic forecasting that encompasses different approaches to obtain distribution forecasts as the output of a NN. The general framework is illustrated in Figure \ref{fig:nn_diagram} and forms the basis of our work here. 
In this section, we briefly introduce three NN variants and refer to \citet{Schulz2022} for details.\footnote{
Note that other types of distributional forecasts such as normalizing flows \citep{Kobyzev2021} exist. 
However, we do not consider them here in the interest of brevity and since the approaches discussed in Section \ref{ssec:prob_networks_distr} share appealing properties with regards to the aggregation methods.
}

While these three variants differ in their characterization of the forecast distribution and the loss function employed in the NN, their use in practice shares a common methodological feature that constitutes the main motivation for our work here. 
As discussed in the introduction, extant practice in NN-based forecasting often relies on DEs. 
This raises the question of how the distribution forecast from the three NN variants can be combined using the aggregation methods described in Section \ref{ssec:agg_methods}, which we will discuss below.

\subsubsection{Distributional regression network (DRN)} \label{ssec:nn_drn}

In the distributional regression network (DRN) approach, the forecasts are issued in the form of a parametric distribution. Under the parametric assumption $F_\theta$, the predictive distribution is characterized by the distribution parameter (vector) $\theta \in \Theta \subset \mathbb{R}^d$, where $\Theta$ is the parameter space. 
Different variants of the DRN approach have been proposed over the past years and can be traced back to at least \citet{Bishop1994}. \cite{Lakshminarayanan2017} and \cite{Rasp2018} use a normal distribution with $\theta = (\mu, \sigma)$, 
\cite{Schulz2022} use a zero-truncated logistic distribution with $\theta = (\mu, \sigma)$, 
where for both distributions $\mu \in \R$ is the location and $\sigma > 0$ the scale parameter, and \citet{Bishop1994} and \citet{DInsantoPolsterer2018} use a mixture of normal distributions. 
To estimate the parameters of the NN, proper scoring rules such as the CRPS \citep{Rasp2018,DInsantoPolsterer2018,Schulz2022} or the negative log-likelihood \citep{Lakshminarayanan2017} serve as custom loss functions.
Extensions of the DRN approach to other parametric families are generally straightforward provided that analytical closed-form expressions of the selected loss function are available \citep[for example,][]{Ghazvinian2021,Chapman2022}.

For VI, distributions from location-scale families form a special case that allows for straightforward aggregation.
Given a CDF $F_{(0)}$, a distribution is said to be an element of a location-scale family if its CDF $F$ satisfies
\begin{eqnarray}
	F ( z; \mu, \sigma ) = F_{(0)} \left( \dfrac{z - \mu}{\sigma} \right), \quad z \in \R,
	\nonumber
\end{eqnarray}
where $\mu \in \R$ denotes the location and $\sigma > 0$ the scale parameter. 
Popular examples include the normal and logistic distributions.
For location-scale families, VI is shape-preserving, which means that if the individual forecasts are elements of the same location-scale family, the aggregated forecast is as well \citep{Thomas1980}. 
Further, the parameters of the aggregated forecast $\mu^\text{VI}$ and $\sigma^\text{VI}$ are given by the weighted averages of the individual parameters $\mu_i$ and $\sigma_i$, $i=1,...,n$, together with the intercept $a$ in case of the location parameter, that is,
\begin{eqnarray}
\mu^\text{VI} = a + \sum_{i=1}^n w_i \mu_i = a + w_0 \sum_{i=1}^n \mu_i, \quad 
\text{and} \quad \sigma^\text{VI} = \sum_{i=1}^n w_i\sigma_i = w_0 \sum_{i=1}^n \sigma_i,
\label{eq:vi_loc_scale}
\end{eqnarray}
where the second equalities each hold under our assumption of equal weights.
% Note that we here only consider the case of $w_i = w_0$ for $i=1,...,n$.
Unlike VI, the LP results in a wide-spread, multi-modal distribution, and is thus not shape-preserving for location-scale families. 
Both \citet{Lakshminarayanan2017} and \citet{Rasp2018} generate DEs based on random initilalization.
While \citet{Lakshminarayanan2017} propose to use the LP to aggregate the forecast distributions, \citet{Rasp2018} instead combine the forecasts by averaging the distribution parameters.
Since the normal distribution
% (which we will also employ in the simulation study below) 
is a location-scale family, parameter averaging is equivalent to \Vo.
For the application in Section \ref{sec:cs}, we will employ a normal distribution, i.e., we obtain the VI forecasts via \eqref{eq:vi_loc_scale}.
To evaluate the LP forecasts, we draw a random sample of size 1,000 from the mixture distribution by first randomly choosing an ensemble member and then generating a random draw from the corresponding distribution.

\subsubsection{Bernstein quantile network (BQN)} \label{ssec:nn_bqn}

\citet{Bremnes2020} proposes a semi-parametric extension of the DRN approach we refer to as  Bernstein quantile network (BQN). The probabilistic forecast is given in form of the quantile function $Q$, which is modeled as a linear combination of Bernstein polynomials, that is,
\begin{eqnarray}
	Q \left( p \right) := \sum_{j = 0}^{d} \alpha_j B_{jd} (p), \quad p \in \left[0, 1\right], \nonumber
\end{eqnarray}
where $\alpha_0 < \dots < \alpha_d$ and $B_{jd}$ is the $j$-th basis Bernstein polynomial of degree $d \in \N$, $j = 0, \dots, d$. 
The basis coefficients $\alpha_0, \dots, \alpha_d$, which define the predictive distribution, are obtained as output of the NN. The parameters of the NN are estimated by minimizing the quantile loss evaluated at pre-defined quantile levels. Note that the support of the forecast distribution is equal to $\left[ \alpha_0, \alpha_d \right]$ and therefore bounded.

To aggregate ensembles of BQN forecasts, \cite{Bremnes2020} and \cite{Schulz2022} average the individual basis coefficient values across ensemble members. 
Resembling the shape-preservation for location-scale families in case of DRN,
this is equivalent to \Vo, which is obvious from the quantile function of the general case of VI for BQN forecasts,
\begin{eqnarray}
	Q_w (p) 
 = a + \sum_{i = 1}^{n} w_i \left( \sum_{j = 0}^{d} \alpha_{ij} B_{jd} (p) \right) 
 % = a + \sum_{j = 0}^{d} \left( \sum_{i = 1}^{n} w_i \alpha_{ij} \right) B_{jd} (p), \quad p \in \left[0, 1\right], \nonumber 
 = \sum_{j = 0}^{d} \left( a + \sum_{i = 1}^{n} w_i \alpha_{ij} \right) B_{jd} (p), \quad p \in \left[0, 1\right], \nonumber 
\end{eqnarray}
where $\alpha_{ij}$ is the coefficient of the $j$-th basis polynomial of the $i$-th ensemble member, $i = 1, \dots, n$, $j = 0, \dots, d$. 
Note that we can move the intercept $a$ into the summation, as the sum of the Bernstein basis polynomials equals 1. Further, we see that we only need to add the intercept to the averaged coefficients to obtain the vincentisized BQN forecast.

Since a closed form of the CDF or density of a BQN forecast is not readily available, the LP cannot be expressed in a similar fashion. Analogous to DRN, the evaluation of the LP forecasts will therefore be based on a random sample of size 1,000 drawn from the aggregated distribution. Here, the inversion method allows to sample from the individual BQN forecasts. Further, the VI forecasts are evaluated based on a sample of 99 equidistant quantiles.\footnote{The numbers of samples and quantiles were chosen based on simulation experiments and theoretical considerations. Compared to random samples from the forecast distributions, a smaller number of equidistant quantiles is required to achieve approximations of the same accuracy, see \citet{KruegerEtAl21} and references therein for a discussion of sample-based estimation of the CRPS.}

\subsubsection{Histogram estimation network (HEN)} \label{ssec:nn_hen}

\begin{figure}
\begin{center}
	\includegraphics[width=\textwidth]{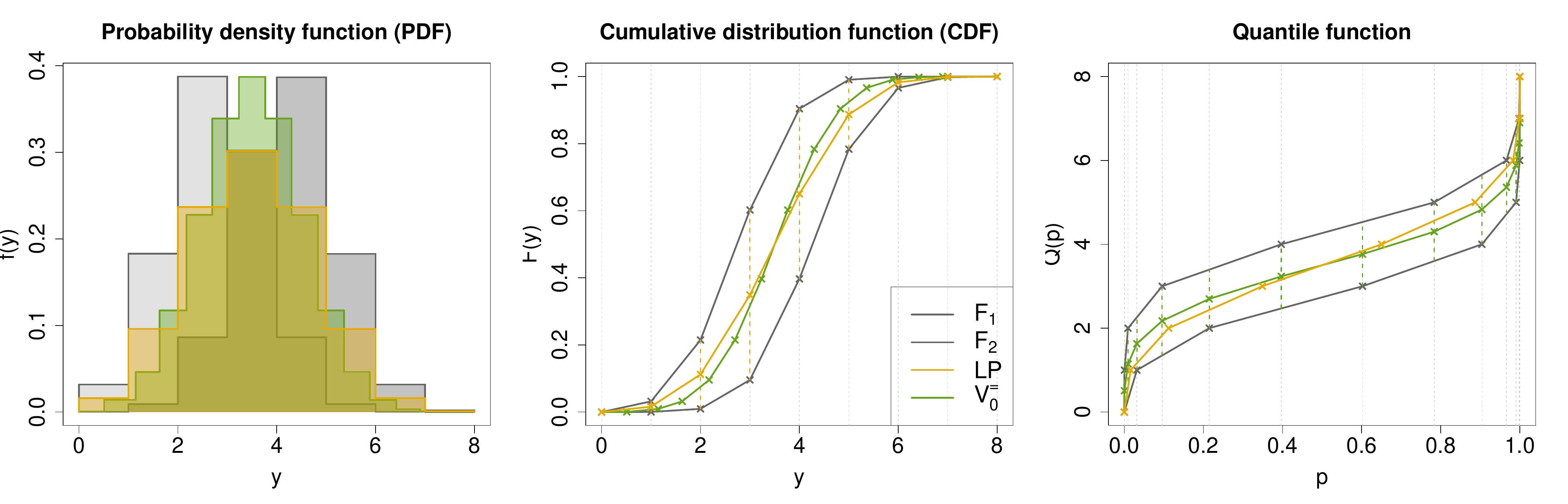}
	\caption{PDF, CDF and quantile function of two HEN forecasts $F_1$ and $F_2$ together with forecasts aggregated via the LP and \Vo. The dashed vertical lines indicate the binning with respect to $F_1$, $F_2$ and $F_w$ for the CDF plot and with respect to $Q_w$ in the quantile function plot. \label{fig:hen_aggregation}}
\end{center}
\end{figure}

The last method considered here is the histogram estimation network (HEN) which divides the support of the target variable in $N \in \N$ bins and assigns each bin the probability for the observation falling in that bin. Variants of this approach have been proposed in a variety of applications \citep[for example,][]{Gasthaus2019,Li2021}. 
Mathematically, the HEN forecast is given by a piecewise uniform distribution. Let $b_0 < \dots < b_N$ denote the edges of the bins $I_\ell = [b_{\ell-1}, b_\ell)$ with probabilities $p_\ell$, $\ell = 1, \dots, N$, where it holds that $\sum_{\ell = 1}^{N} p_\ell = 1$. The CDF of a HEN forecast is then given by the piecewise linear function
\begin{eqnarray}
	F (z) = \sum_{\ell = 1}^{N} \left( p_\ell \cdot \dfrac{ \tilde{z} - b_{\ell-1} }{ b_\ell - b_{\ell-1} } \cdot \mathbbm{1} \lbrace b_{\ell-1} \leq z \rbrace \right) \; \text{ with } \; \tilde{z} := \max \left( b_{\ell-1}, \min \left( b_\ell, z \right) \right), \quad z \in \R. \nonumber
\end{eqnarray}
Note that a piecewise linear CDF corresponds to a piecewise linear quantile function and a piecewise constant PDF that resembles a histogram. Figure \ref{fig:hen_aggregation} illustrates the shape of these functions for exemplary HEN forecasts.

We here follow \cite{Schulz2022} by considering fixed bins and estimate only the corresponding probabilities as output of the NN. 
In contrast, we here standardize the target variable and define the bin edges based on quantiles of the standard normal distribution. For prediction (and evaluation), the bin edges can easily be transformed back to the original scale of the target variable.
As for DRN, the NN can be trained via CRPS minimization or maximum likelihood. Here, we use the latter, which corresponds to minimizing the categorical cross-entropy, a standard approach for classification tasks in machine learning.

Regarding the aggregation of HEN forecasts in case of fixed bins, the LP is equivalent to averaging the bin probabilities since
\begin{eqnarray}
	 F_w (z) = \sum_{i = 1}^{n} w_i \left[ \sum_{\ell = 1}^{N} \left( p_{i\ell} \dfrac{ \tilde{z} - b_{\ell-1} }{ b_\ell - b_{\ell-1} } \mathbbm{1} \lbrace b_{\ell-1} \leq z \rbrace \right) \right] = \sum_{\ell = 1}^{N} \left[ \left( \sum_{i = 1}^{n} w_i p_{i\ell} \right) \dfrac{ \tilde{z} - b_{\ell-1} }{ b_\ell - b_{\ell-1} } \mathbbm{1} \lbrace b_{\ell-1} \leq z \rbrace \right], \nonumber 
\end{eqnarray}
where $z \in \R$ and $p_{i\ell}$ is the probability of the $\ell$-th bin for the $i$-th ensemble member, $i = 1, \dots, n$, $\ell = 1, \dots, N$. An exemplary application of the LP for an approach akin to HEN forecasts in a stacked NN can be found in \cite{Clare2021}.
By contrast to the LP, the VI approach exhibits a particular advantage for HEN forecasts in that it results in a finer binning than the individual HEN models. To illustrate this effect, we note that the quantile function is a piecewise linear function with edges depending on the accumulated bin probabilities, 
that is, $p^{*}_\ell := \sum_{m = 1}^{\ell} p_m$, $\ell = 1, \dots, N$.
In mathematical terms, the quantile function is given for $p \in \left[ 0 , 1 \right]$ by 
\begin{eqnarray}
	Q (p) = b_0 + \sum_{\ell = 1}^{N} \left( b_\ell - b_{\ell-1} \right) \left( \dfrac{ \tilde{p} - p^{*}_{\ell-1} }{ p^{*}_\ell - p^{*}_{\ell-1} } \cdot \mathbbm{1} \lbrace p^{*}_{\ell-1} \leq p \rbrace \right) \; \text{ with } \; \tilde{p} := \max \left( p^{*}_{\ell-1}, \min \left( p^{*}_{\ell}, p \right) \right). \nonumber
\end{eqnarray}
Therefore, the resulting VI quantile function is a piecewise linear function with one edge for each accumulated probability of the individual forecasts. 
As the forecast probabilities differ for each member of the DE, the associated quantile functions are subject to a different binning. 
Since the set of edges of the aggregated VI forecast is given by the union of all individual edges, this leads to a smoothed final forecast distribution with a finer binning than the individual model runs that differs for every forecast case, and eliminates the potential downside of too coarse fixed bin edges. 
Figure \ref{fig:hen_aggregation} illustrates the effects of the LP and \Vo\ for two exemplary HEN forecasts, where the binning of the \Vo\ forecast distribution is finer than that of the individual forecasts and of the LP.

\subsection{Ensembling strategies} \label{ssec:nn_ens}

Various methods have been proposed for the generation of DEs each addressing different aspects of uncertainty in the training process. For this study, we picked the most common DE approaches, the underlying ideas of which we briefly present in the following. Implementation details are deferred to Appendix \ref{supl:nn_setup}.

\subsubsection*{Naive ensemble} \label{sssec:nn_naive}

Due to the random initialization of the NN weights and the stochastic gradient descent algorithm, the process of training a standard NN is subject to stochasticity and therefore multiple training runs will result in different weight estimates. Hence, a straightforward way to generate a DE is to simply train several models based on different random initializations, which we refer to as naive ensemble. It is not only simple to implement, but has also been shown to result in improved predictive performance \citep[see, e.g.,][]{Lakshminarayanan2017,Fort2019}.

\subsubsection*{Bagging} \label{sssec:nn_bagging}

Bagging (“bootstrap aggregating”) is one of the earliest ideas for generating ensembles \citep{Breiman1996} and forms the basis of other ensembling-based methods such as random forests \citep{Breiman2001}. Bagging generates multiple models such that each is based on a different bootstrapped sample of the original training data. For NNs, the ensemble models thus do not only differ due to the random initialization and stochastic gradient descent, but also due to the bootstrapped training sets.

\subsubsection*{BatchEnsemble} \label{sssec:nn_batchens}

Although parallelizable, one disadvantage of the naive ensemble and bagging is that the computational costs increase linearly with the ensemble size as no modifications are applied to make the ensemble generation more efficient. To address this, \citet{Wen2020} introduce BatchEnsemble, an efficient ensemble method with parallel mini-batch training and shared weights, which reduces the computational costs significantly and performs comparably to the naive ensemble in their original study.

\subsubsection*{Dropout variants (MC dropout, variational dropout, concrete dropout)} \label{sssec:nn_drop}

A widely used technique for regularization, that can also be used for ensembling, is dropout \citep{Srivastava2014}, which operates by randomly dropping neurons with a given probability. 
We consider three variants of dropout that differ in the choice of the dropout rate: Monte Carlo (MC) dropout treats the (overall) dropout rate as additional hyperparameter that is chosen beforehand, variational dropout learns the dropout rate during training \citep{Kingma2015}, and concrete dropout improves over the former by adapting the rate-learning process allowing that the rate is learned directly per layer \citep{Gal2017}. 
Here, we apply dropout both during training and inference, where we generate ensembles by repeatedly predicting with one base model \citep{GalGhahramani2016},
a mechanism that inherently differs from the previous strategies. 
In the following, we will differentiate between multi- and base-model approaches.

\subsubsection*{Bayesian Neural Networks} \label{sssec:nn_bayesian}

The final method for ensemble generation is based on Bayesian neural networks \citep[BNNs;][]{Lampinen2001,Jospin2022}, which account for the uncertainty in the learning task using Bayesian ideas. Instead of learning the weights of the NN as deterministic values, the distribution of the weights is modelled. By sampling from the learned distributions, we can generate ensembles of predictions. 
As for the dropout models, we train one base model that is used for the generation of the DE, i.e., this strategy is also a base-model approach.

\section{Case study} \label{sec:cs}

We compare the performance of the five aggregation methods for each of the three NN variants and seven ensembling strategies on various data sets, which comprise a data set on wind speed forecasting \citep{Schulz2024dataset}, two simulated data sets \citep{Li2021} and nine open-source machine learning benchmark data sets \citep[see, e.g.,][]{GalGhahramani2016,Lakshminarayanan2017}.\footnote{
An earlier version of the manuscript only included the two simulation studies and the wind gust data. Following the inclusion of the benchmark data sets, we decided to keep the wind data and simulations to have a larger variety of data sets.
} 

As a detailed analysis for all combinations of data sets, ensembling strategies and NN variants is too cumbersome, we first provide an in-depth analysis for two selected cases and then investigate results over all combinations on a higher level. The in-depth analysis is carried out to highlight reoccurring effects of the aggregation methods on the predictive performance in terms of scores, calibration and sharpness.
While we observe similar effects within the application to certain data sets and ensembling strategies, the properties of DE and aggregation differ for each data set and ensembling strategy. Hence, we also investigate the aggregation methods over all combinations, where we draw conclusions on the effects of aggregation based on the underlying characteristics of the DE that differ over the cases.

To account for the uncertainty in data sampling, we generate different partitions of each data set by splitting them multiple times in training, validation and test sets. We follow \citet{GalGhahramani2016} and use 20 random partitions of the data sets except for the Protein and Wind data sets, where the number is restricted to 5 due to the size. For the 9 machine learning benchmark data sets, the training, validation and test set each make up 72\%, 18\% and 10\%, respectively. For the 5 partitions of the Wind data set, we use one of the calendar years as test set for each of the 5 partitions and 20\% of the remaining data for validation, as for the other data sets. The two synthetic data sets, which are referred to as Scenarios 1 and 2, are adopted from \citet{Li2021} and described in more detail in Appendix \ref{supl:sim_studies}. For Scenarios 1 and 2, we do not create partitions from the same data set but instead by repeated random generation. 
For each partition, we then calculate 20 ensemble members, which are used to build ensembles of sizes 2, 4, \dots, 20. In the interest of computational requirements, we use steps of size 2 and restrict the maximum ensemble size to 20. Further, previous tests have shown that increasing the ensembles above a size of 20 results only in a marginal improvement in the performance.
A summary of the data is provided in Table \ref{tbl:data_sets}.

\begin{table}
\caption{Overview of the data set sizes. 
\label{tbl:data_sets}}	
\begin{center}
\begin{tabular}{l@{\hskip 0.5cm}r@{\hskip 0.5cm}r@{\hskip 0.75cm}r@{\hskip 0.75cm}r@{\hskip 0.75cm}r@{\hskip 0.75cm}l}
	\toprule 
	Data set & Total & Training & Validation & Testing & Features \\
	\midrule
	Wind & 378,833 & 252,946 & 63,237 & 62,650 & 67 \\ 
	\midrule
	Scenarios 1--2 & 7,000 & 5,000 & 1,000 & 1,000 & 5 \\
	\midrule
	Protein & 45,730 & 32,925 & 8,232 & 4,573 & 9 \\
	Naval & 11,934 & 8,592 & 2,149 & 1,193 & 16 \\
	Power & 9,568 & 6,888 & 1,723 & 957 & 4 \\
	Kin8nm & 8,192 & 5,898 & 1,475 & 819 & 8 \\
	Wine & 1,599 & 1,151 & 288 & 160 & 11 \\
	Concrete & 1,030 & 741 & 186 & 103 & 8 \\
	Energy & 768 & 552 & 139 & 77 & 8 \\
	Boston & 506 & 364 & 91 & 51 & 13 \\
	Yacht & 308 & 222 & 55 & 31 & 6 \\
	\bottomrule
\end{tabular}
\end{center}
\end{table}

For each combination of data set, ensembling strategy and NN variant, we perform hyperparameter tuning and choose one combination of hyperparameters, which is then used for all partitions. Over a set of pre-defined choices,
we chose the best-performing models on the validation sets of the first two random partitions. More details on the tuning procedure and the chosen hyperparameters are provided in Appendix \ref{supl:hpar_tuning}. 
For evaluation, we will assess the improvement from aggregation by comparison with the corresponding average of the DE. Hence, the CRPSS in \eqref{eq:crpss} will be calculated using the average CRPS of the individual NNs for $\bar S_{\text{ref}}$.\footnote{
	Note that this does not correspond to the mean improvement over the individual forecasts. However, averaging the median skill scores of the individual ensemble member predictions over the repetitions of the simulations yields qualitatively analogous results (not shown).
}
If not noted otherwise, the evaluation is based on the mean values computed for each combination of the factors in Table \ref{tbl:cs_factors}, in particular, we also calculate the mean value for each partition separately.
Table \ref{tbl:cs_factors} provides an overview of all the factors in Section \ref{ssec:evaluation}.

\begin{table}
\caption{Design parameters for the application of the aggregation of DEs in Section \ref{sec:cs}. 
In total, we obtain 88,200 forecast configurations for DEs and 220,500 for aggregation.
\label{tbl:cs_factors}}	
\begin{center}
\begin{tabular}{l@{\hskip 0.5cm}r@{\hskip 0.5cm}l@{\hskip 0.3cm}}
	\toprule 
	Factor & Size & Values \\
    \midrule
	Forecast distribution & 3 & DRN, BQN, HEN \\
	\midrule
	Aggregation method & 5 & LP, \Vo, \Va, \Vw, \Vaw\ \\
	\midrule
	Ensembling strategy & 7 & Naive ensemble, bagging, BatchEnsemble, MC dropout, \\
	&  & variational dropout, concrete dropout, Bayesian NN \\
	\midrule
	Data set & 12 & Wind, Scenarios 1--2, Protein, Naval, Power, Kin8nm, \\
	& & Wine, Concrete, Energy, Boston, Yacht \\
	\midrule
	Ensemble size & 10 & 2, 4, ..., 20 \\
	\midrule
	Partition & 5/20 & 1, 2, ..., 5 for Wind and Protein, 1, 2, ..., 20 otherwise \\
	\bottomrule
% \end{tabularx}
\end{tabular}
\end{center}
\end{table}

\subsection{In-depth analysis} \label{ssec:cs_indepth}

For the in-depth analysis, we pick two cases that highlight potential effects of aggregation on DE forecasts of the three NN variants. The two cases have been chosen as they are typical for reoccurring situations in which certain patterns arise.
The first example is based on DEs generated with Bayesian NNs and the Kin8nm data set, which has a size of 8,192. The second example is based on another ensemble strategy and a much smaller data set with only 506 samples, namely, Bagging and the Boston data set. 

\subsubsection*{Kin8nm and Bayesian deep ensembles}

\begin{figure}
	\begin{center}
		\includegraphics[width=\textwidth]{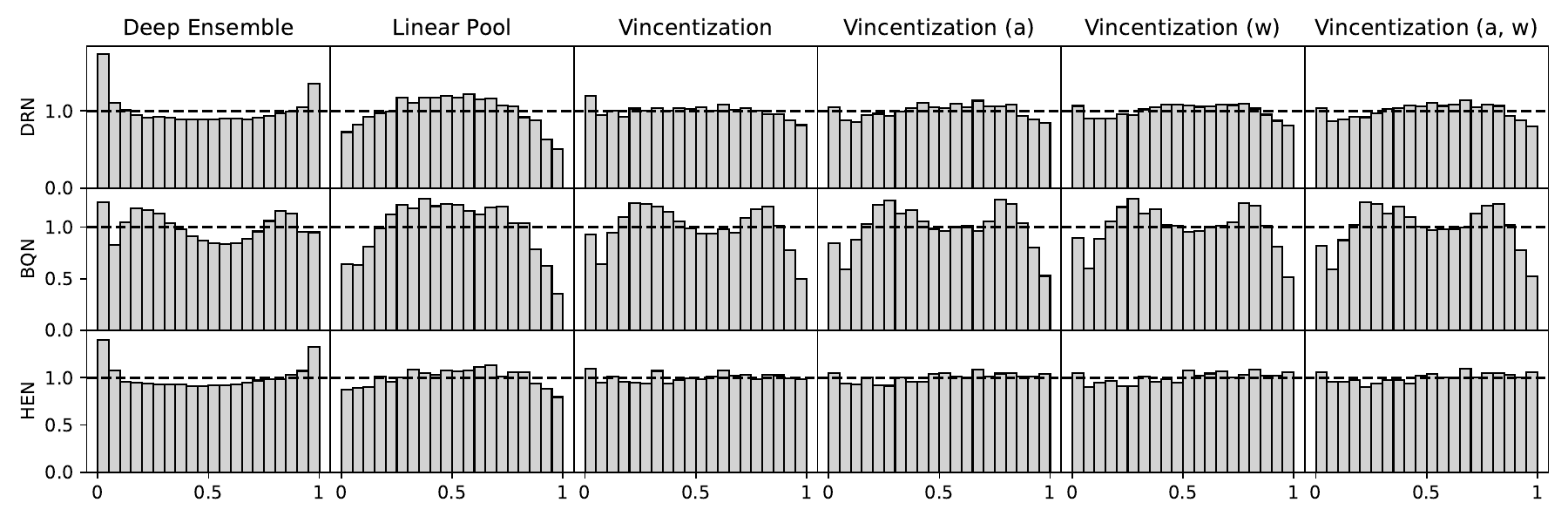}
		\caption{PIT histograms of Bayesian DEs and the aggregation methods for the three NN variants and the Kin8nm data. The ensembles are of size 10.
  \label{fig:kin8nm_upit} }
\end{center}
\end{figure}

First, we visually inspect the calibration of both the DE forecasts and the aggregated forecasts via the PIT histograms in Figure \ref{fig:kin8nm_upit}. We find that none of the NN variants generate calibrated forecasts. Both the DRN and HEN forecasts are underdispersive resulting in an U-shaped histogram, while the BQN forecasts result in a wave-shaped histogram. Comparing with the aggregated forecasts, we find that the shapes change systematically. For the LP, we obtain forecasts that are overdispersed but to a different extent for all three NN variants. This effect is also observed for all other cases, as it aligns with the theoretical property that the LP increases the dispersion with respect to the individual ensemble members. All VI forecasts of DRN and HEN result in flat histograms indicating calibrated forecasts. For BQN, the VI forecasts have the same wave-like shape as the individual forecasts and seem to be a bit overdispersed. Hence, they were not able to correct for this specific kind of miscalibration. In general, miscalibration different from the typical under- and overdispersion are not corrected by aggregation. Between the VI variants, we do not observe systematic differences.

\begin{figure}
\begin{center}
		\includegraphics[width=\textwidth]{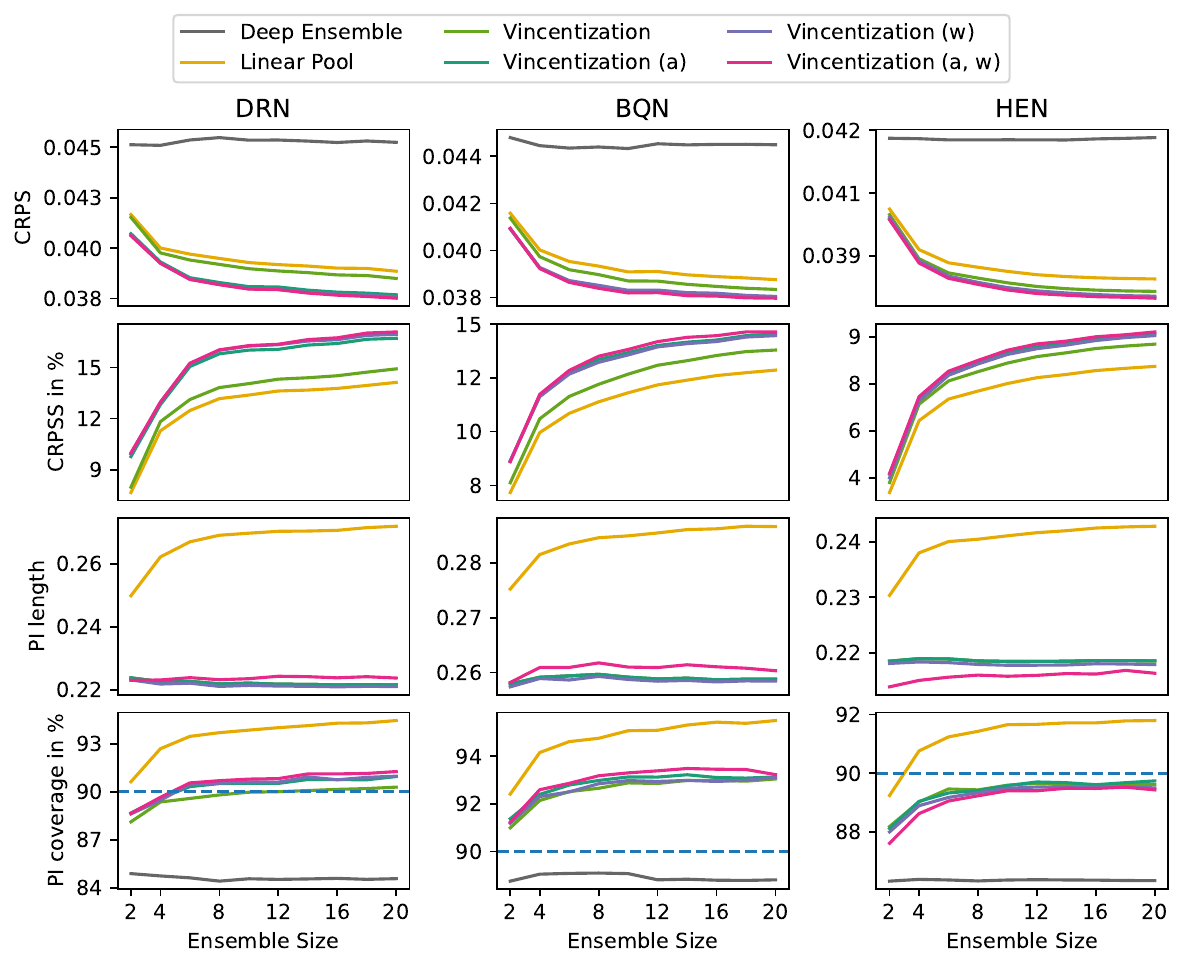}
		\caption{Evaluation metrics of Bayesian DEs and the aggregation methods for the three NN variants and the Kin8nm data. Note the different scales on the vertical axis. \label{fig:kin8nm_scores}}
\end{center}
\end{figure}

Following the visual inspection of calibration, we quantitatively analyze the predictive performance dependent on the size of the ensemble via Figure \ref{fig:kin8nm_scores}. Unsurprisingly, all aggregation methods improve upon the individual forecasts in terms of the CRPS, for each NN variant to a different extent. The ranking is identical over the different NN variants with the VI approaches using parameter estimation performing best, followed by \Vo\ and LP, a ranking typical for base-model approaches. 
Most improvement from increasing the ensemble size is obtained up to ensembles of size 10, a pattern that will reemerge in the overall analysis. 
Looking at the PI length, we find that while \Vw\ and \Vaw\ have only a small influence on the PI length, the LP increases the PI length by a larger margin explaining the increase in dispersion observed in the PIT histograms. 
Note that \Vo\ and \Va\ do not affect the PI length by definition and are therefore not included in the analysis of the PI lengths. 
Further, the PI length of the LP increases strongly for small ensemble sizes and remains almost constant for larger sizes. 
This increase in the PI length of the LP is resembled in the corresponding coverage, where the LP yields the PIs with the largest coverages, much larger than that of the individual forecasts. The VI forecasts of DRN and HEN are not only calibrated, but also their associated coverages are close to the nominal level. For BQN, all aggregation methods increase the coverages beyond the nominal level deviating even more. 
The increase in PI coverage by aggregation is also observed in most of other cases.

\begin{figure}
	\begin{center}
	\includegraphics[width=\textwidth]{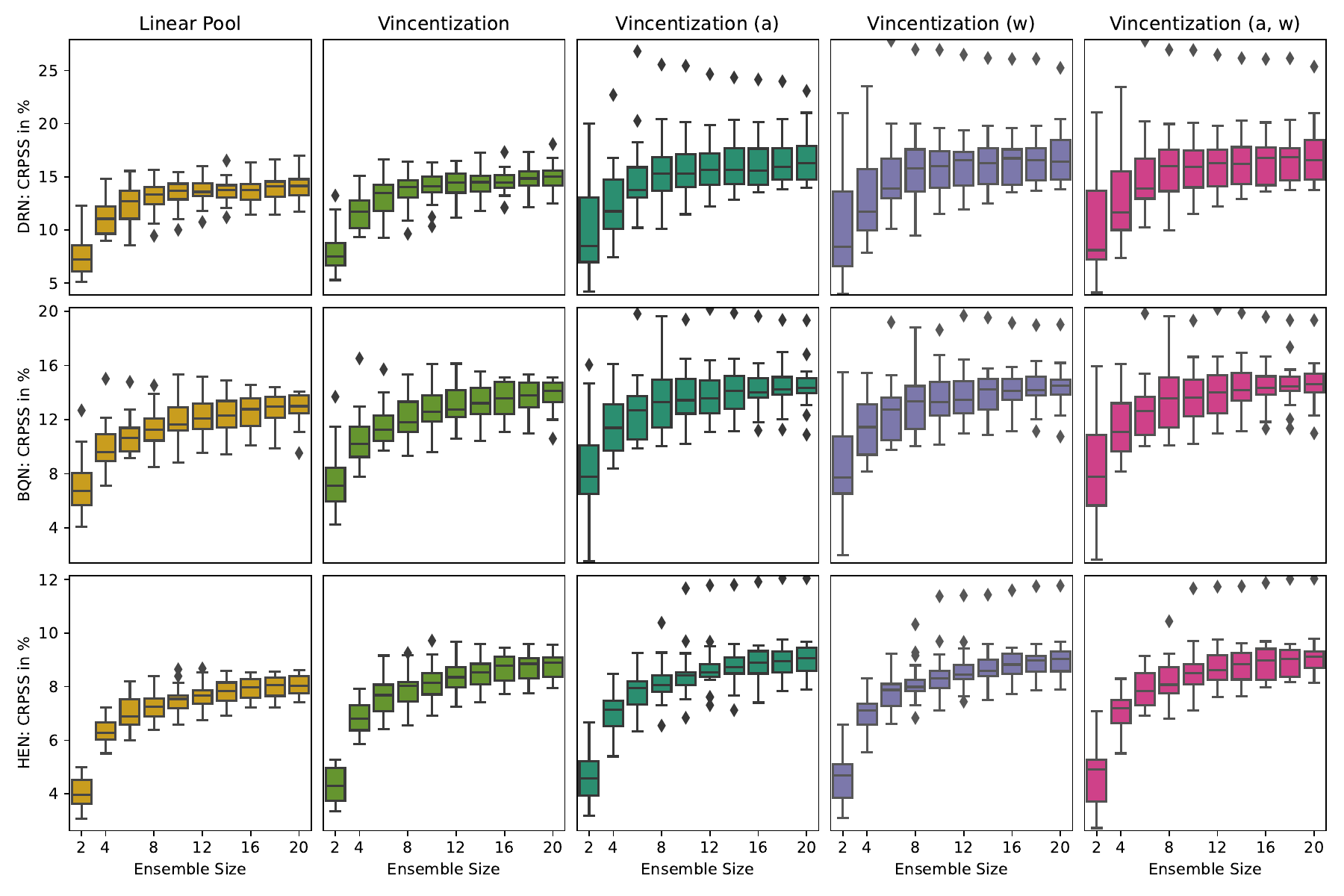}
	\caption{Boxplots over the CRPSS values of the aggregation methods for each ensemble size for Bayesian DEs and the Kin8nm data. 
    \label{fig:kin8nm_crpss_boxplots}}
\end{center}
\end{figure}

At last, we analyze the effect of the ensemble size and the variability over the partitions based on Figure \ref{fig:kin8nm_crpss_boxplots}.
% Figure \ref{fig:kin8nm_crpss_boxplots} shows boxplots of the CRPSS over the runs dependent on the ensemble size. 
First, we note that in none of the cases aggregation degrades performance. Also, the boxes seem to stabilize and the variability becomes smaller with increasing ensemble size. Between the aggregation methods, we see that the variants with parameter estimation have a larger variability than LP and \Vo.
Although parameter estimation improves the predictive performance over all partitions, we see that in certain cases, especially small ensemble sizes, it results in the worst performance among the aggregation methods. Contrarily, the best results are also obtained by parameter estimation, i.e., we observe both positive and negative outliers in terms of the performance.
These result are also representative for the remaining cases.

\subsubsection*{Boston and Bagging deep ensembles}

Now, we move on to the Boston data set and Bagging DEs. In contrast to the Kin8nm data set, the Boston data set is relatively small, including only a total of 506 samples.
Another difference to the previous example is that the DEs are not generated using a base-model approach but instead a multi-model approach.

\begin{figure}
	\begin{center}
		\includegraphics[width=\textwidth]{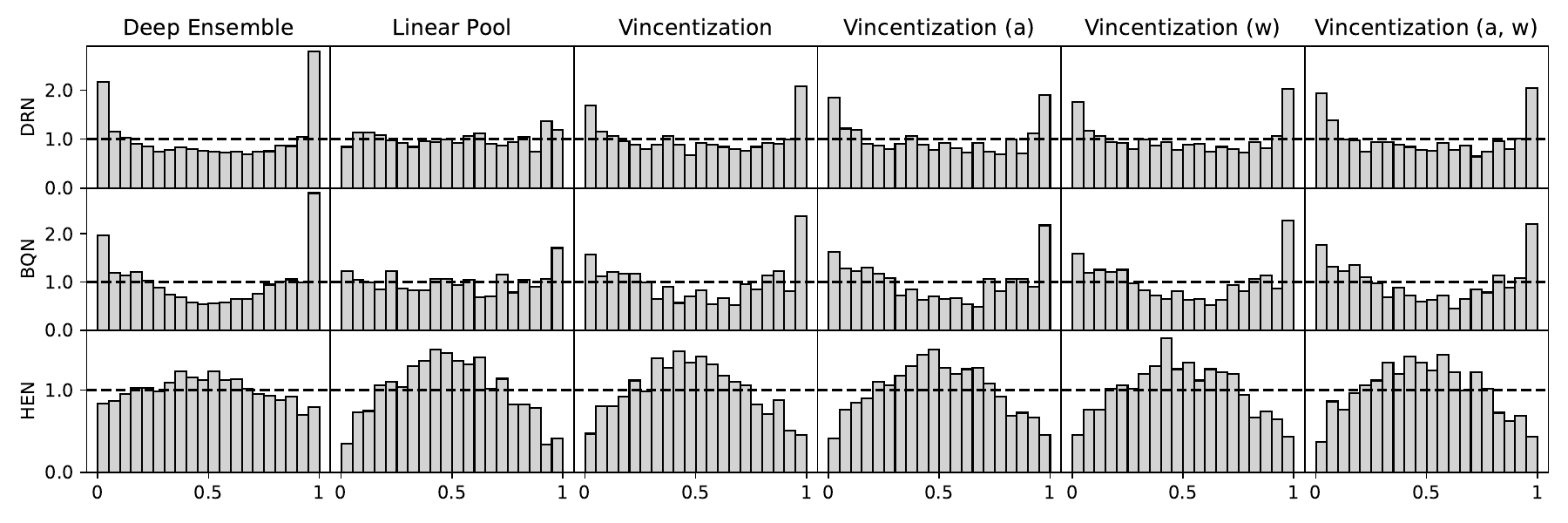}
		\caption{PIT histograms of Bagging DEs and the aggregation methods for the three NN variants and the Boston data. The ensembles are of size 10.
  \label{fig:boston_upit} }
\end{center}
\end{figure}

Again, we start by looking at the PIT histograms in Figure \ref{fig:boston_upit}. While DRN and BQN generate strongly underdispersed forecasts, HEN results in overdispersed forecasts. 
For Bagging, the methods generally tend to result in more underdispersed forecasts.
The LP increases dispersion with respect to the DE, which results in calibrated forecasts for DRN and BQN. This case provides a good example of the strength of the LP for underdispersed resp.\ overconfident forecasts, which are often observed for NNs trained on small data sets, as in our case study. 
In contrast, the VI forecasts are not able to fully correct for the underdispersion but instead only slightly.
For HEN, both the LP and VI forecasts are more overdispersed after aggregation. 
While performance metrics such as the CRPS improve by calibration, there is no guarantee that an aggregation method will improve the calibration of the forecasts, as now seen in both examples.

\begin{figure}
	\begin{center}
		\includegraphics[width=\textwidth]{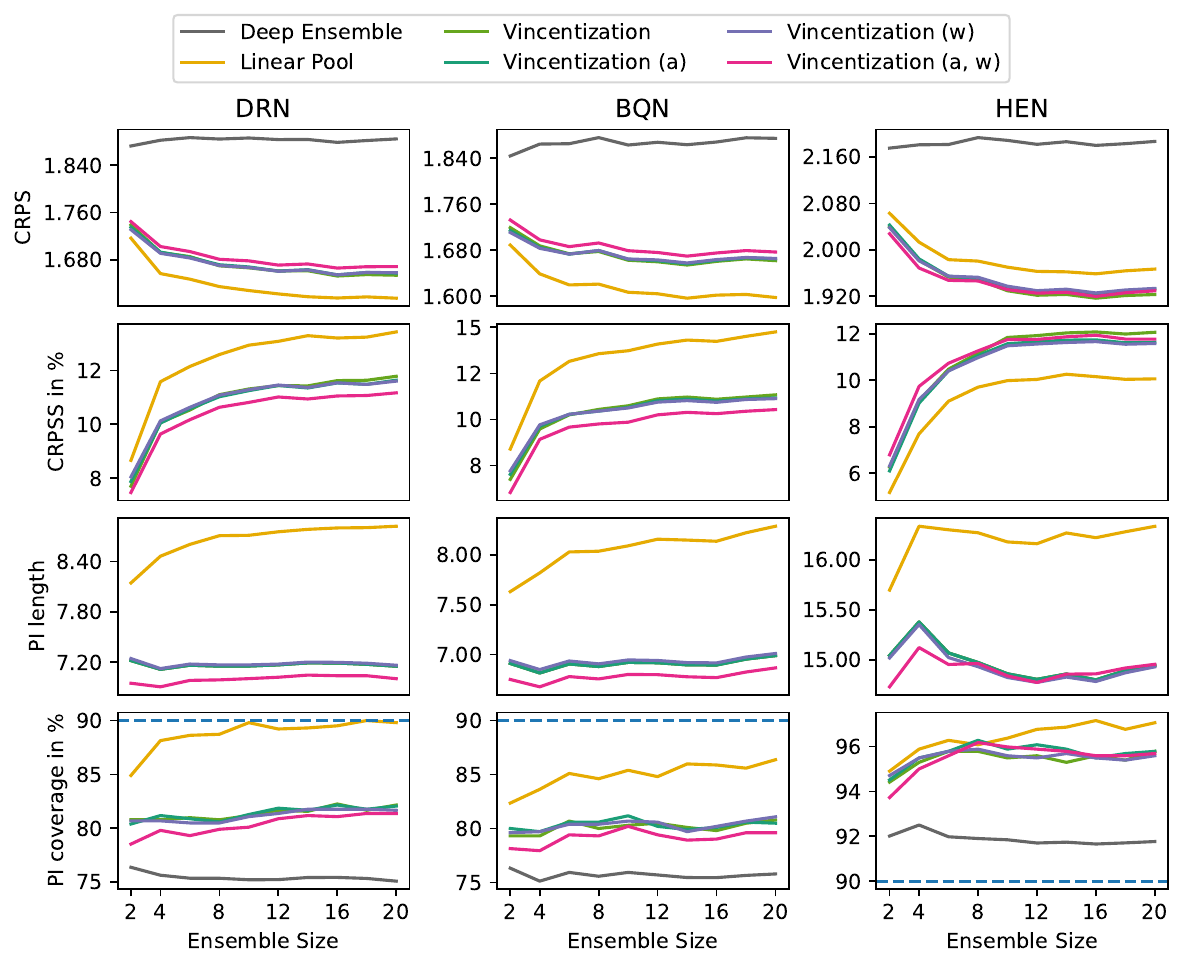}
		\caption{Evaluation metrics of Bagging DEs and the aggregation methods for the three NN variants and the Boston data. Note the different scales on the vertical axis. \label{fig:boston_scores}}
\end{center}
\end{figure}

Turning to the evaluation measures in Figure \ref{fig:boston_scores}, we obtain results that differ from those of the Kin8nm data. 
The LP performs best for DRN and BQN, the two cases for which it generated calibrated forecasts. In case of the VI variants, \Vaw\ has the lowest CRPSS. 
Even though we observed the VI variants performing best for underdispersed forecasts in the Kin8nm example, the LP performs especially well in these situations. Further, base-model approaches generally result in better performance using VI, the effect is not as strong for multi-model approaches such as Bagging.
In contrast, the VI variants outperform the LP for HEN, where we do not observe under- but instead overdispersion. HEN favors aggregation by VI over the LP, also due to the fact that HEN more often results in overdispersed forecasts.
For the PI related measures, we obtain similar results as for the Kin8nm data, i.e., the LP increases the PI lengths drastically and all aggregation methods result in a larger PI coverage than that of the DE. Interestingly, this also holds for \Vaw, which decreases the PI length and results in sharper forecasts despite the observed overdispersion, which might be a result of overfitting, as the validation sets have an average size of 91 samples. Although the PI length becomes smaller, the PI coverage increases. 

Since the results are similar to those observed for the Kin8nm data set, we do not show the effect of the ensemble size and the variability over the partitions analogously to Figure \ref{fig:kin8nm_crpss_boxplots}.
Overall, we conclude that the results of the in-depth analysis agree with the theoretical properties presented in Section \ref{ssec:agg_methods}. 
No aggregation method was superior throughout all cases, in some cases aggregation calibrated the forecasts, in some it did not and increased dispersion even further. Still, aggregation improved the predictive performance with respect to the DE throughout all cases.

\subsection{Comprehensive analysis of all data sets} \label{ssec:cs_bench}

\begin{figure}
	\begin{center}
		\includegraphics[width=\textwidth]{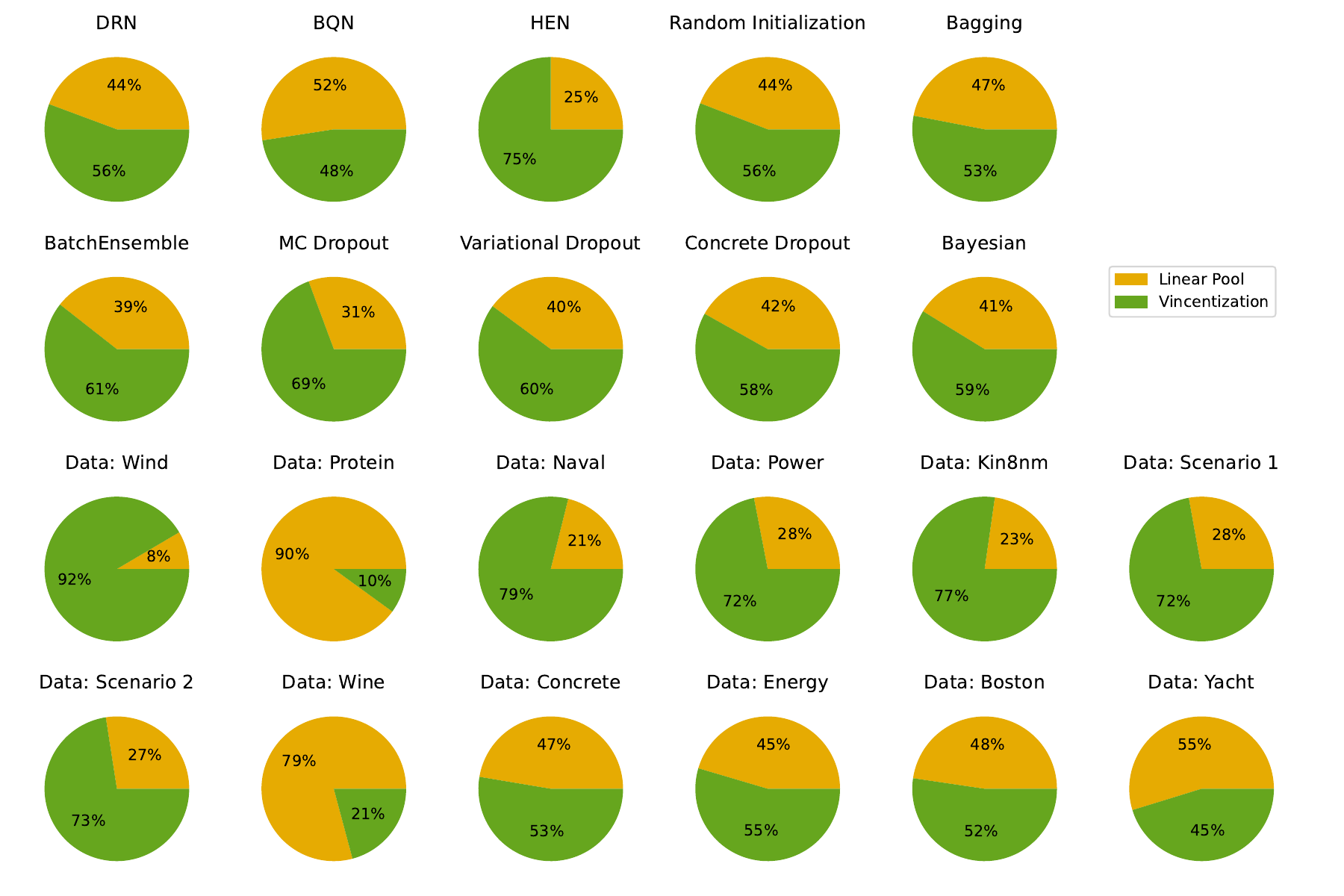}
		\caption{Pie charts showing the proportion of cases in which either the LP is superior to \Vo\ (yellow) or vice versa (green) in terms of the CRPS dependent on the NN variant, ensembling strategy or data set. The data sets are ordered according to their size starting with the largest. \label{fig:bench_pie_vi_lp}}
\end{center}
\end{figure}

Following the detailed analysis of selected examples, we apply the aggregation methods to forecasts of data sets. 
We start the evaluation with an overall analysis of the relative performance of the aggregation methods based on the CRPS. First, we compare only the two aggregation methods that do not require parameter estimation, namely, the LP and \Vo. Figure \ref{fig:bench_pie_vi_lp} shows the proportion of cases where one of the two methods is superior, meaning it has a lower CRPS, dependent on either the NN variant, the ensembling strategy or the data set. While the LP and \Vo\ perform almost equally for DRN and BQN, \Vo\ performs better than LP in three out of four cases for HEN. In case of the ensembling strategies, there is a trend towards \Vo\ with larger proportions for ensembles generated with one base model such as MC dropout. Regarding the data sets, there are two sets for which the LP is the dominant aggregation method, namely, the Protein and Wine data sets. Besides these, \Vo\ is preferred among all data sets but the smallest.
In terms of the size of the data sets, we find that the the proportion of superior \Vo\ cases increases with the size.

\begin{figure}
	\begin{center}
		\includegraphics[width=\textwidth]{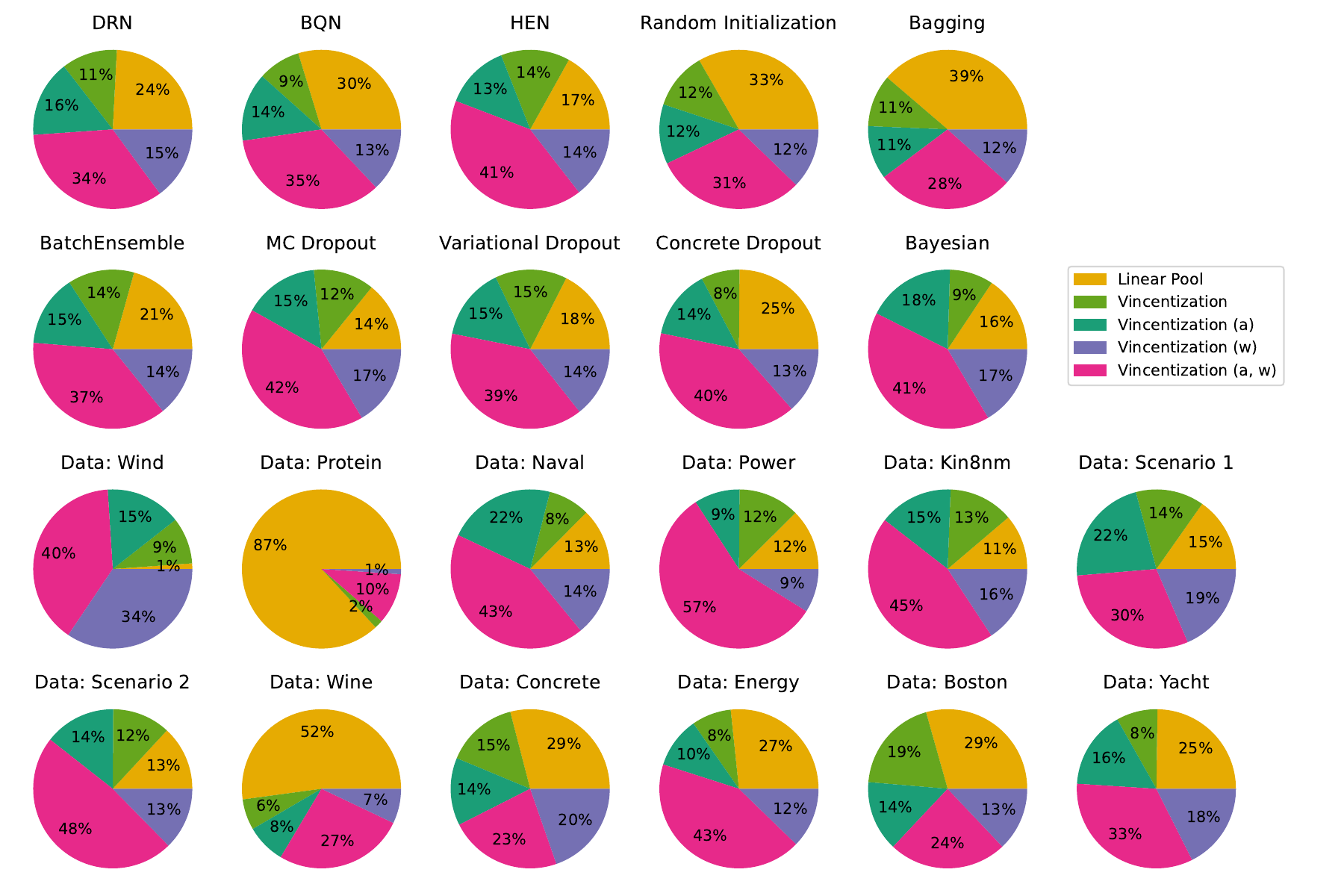}
		\caption{Pie charts showing the proportion of cases in which each of the aggregation methods is superior in terms of the CRPS dependent on the NN variant, ensembling strategy or data set. The data sets are ordered according to their size starting with the largest. \label{fig:bench_pie}}
\end{center}
\end{figure}

If we include the other VI variants with parameter estimation in the comparison, we find that parameter estimation is able to improve upon \Vo, and that the patterns in the differences between LP and \Vo\ persist and even become stronger. The conclusions drawn from the comparison of the LP and \Vo\ can be extended towards the other VI variants. 
In particular, Figure \ref{fig:bench_pie} shows that the VI variants have the largest proportions of best performances for all cases but the Protein and Wine dataset. The VI variants are especially dominant for HEN, the base-model strategies and the larger data sets up to Scenario 2, where the proportions of the VI methods increases with the data set size. 
Among the VI variants, \Vaw\ most often performs best followed by \Va\ and \Vw. This effect also becomes smaller as the sample size decreases. Further, parameter estimation is especially favorable in case of the base-model approaches such as the dropout variants and Bayesian NNs. 
Figure \ref{fig:bench_rank_barplots} in Appendix \ref{supl:figures} shows shows not only the distribution of the best method but instead all ranks. Most interestingly, we find that the LP either performs best or worst most of the time. Hence, we find a clear distinction between LP and VI variants. 

\begin{figure}
	\begin{center}
		\includegraphics[width=0.49\textwidth]{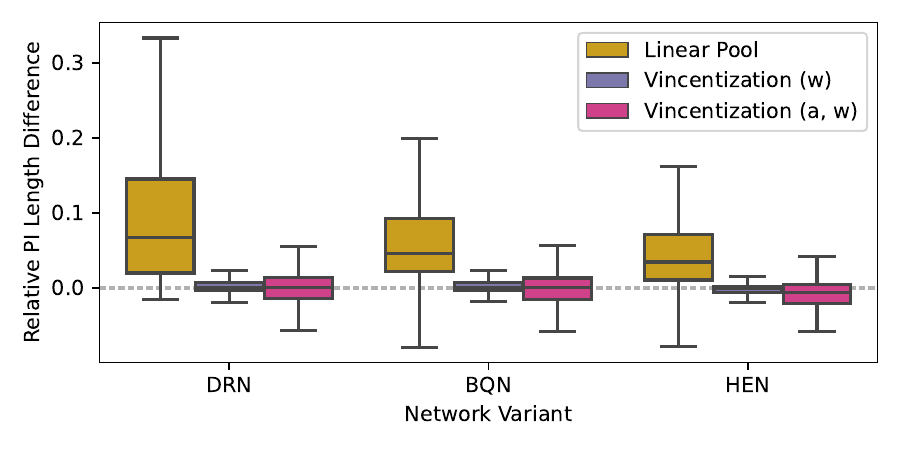}
		\includegraphics[width=0.49\textwidth]{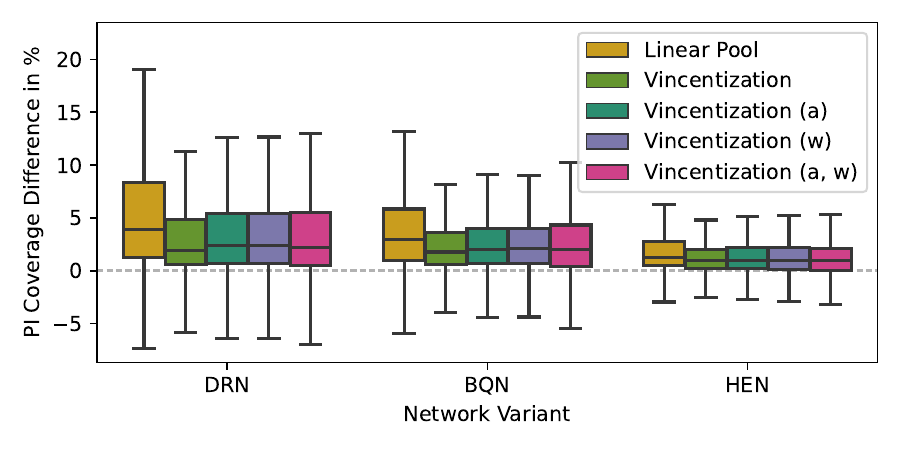}
		\caption{Boxplots of the relative PI length differences (left) and the PI coverage differences (right) with respect to the DE dependent on the aggregation method and NN variant. \label{fig:bench_boxes_lgt_cov_nn}}
\end{center}
\end{figure}

Before we investigate on the effect of the DE characteristics on the performance of the aggregation methods, we briefly analyze the PI lengths. The left panel in Figure \ref{fig:bench_boxes_lgt_cov_nn} shows the relative PI length difference of the aggregated forecasts with respect to associated DE. As expected, we find that the LP increases the PI length in almost all of the cases, while \Vw\ and \Vaw\ are centered around zero. For HEN, \Vaw\ mostly decreases the PI length, which might be a reason that LP does not work as well as VI for this NN variant. As pointed out in the in-depth analyses, the PI length is strongly connected to the PI coverage, which is illustrated analogously in the right panel of Figure \ref{fig:bench_boxes_lgt_cov_nn}. Aligning with previous results, the PI coverage increases in a majority of the cases as all lower quartiles are positive. Figures \ref{fig:bench_boxes_lgt_cov_ens} and \ref{fig:bench_boxes_lgt_cov_datasets} in Appendix \ref{supl:figures} show similar plots dependent on the ensembling strategy and data set. Notably, \Vaw\ results in larger relative PI differences for ensembling strategies that rely on one base model, which might be a reason why \Vaw\ performs particularly well in these cases. For the data sets, the size has again an influence on the results as the differences become in general larger the smaller the data sets become.

\begin{figure}
	\begin{center}
		\includegraphics[width=0.49\textwidth]{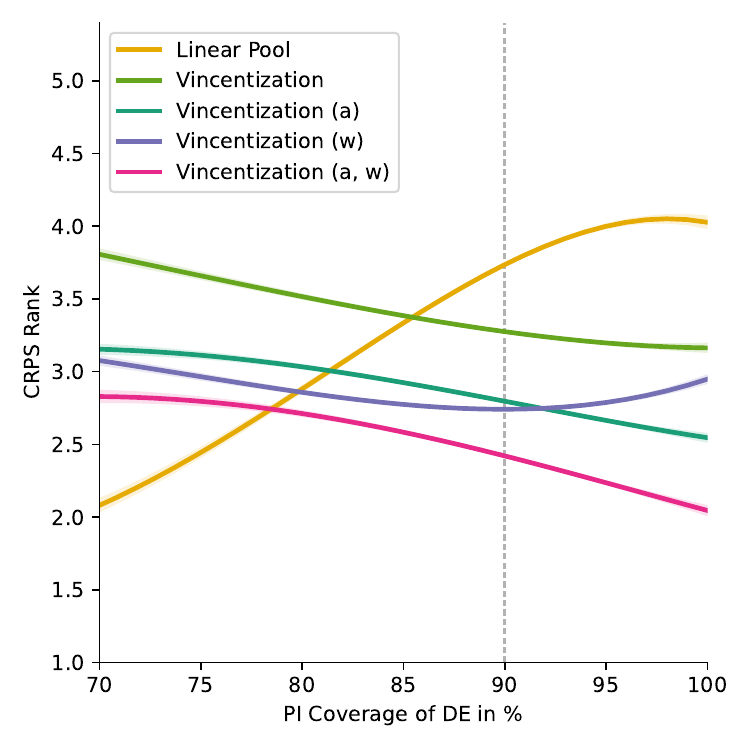}
		\includegraphics[width=0.49\textwidth]{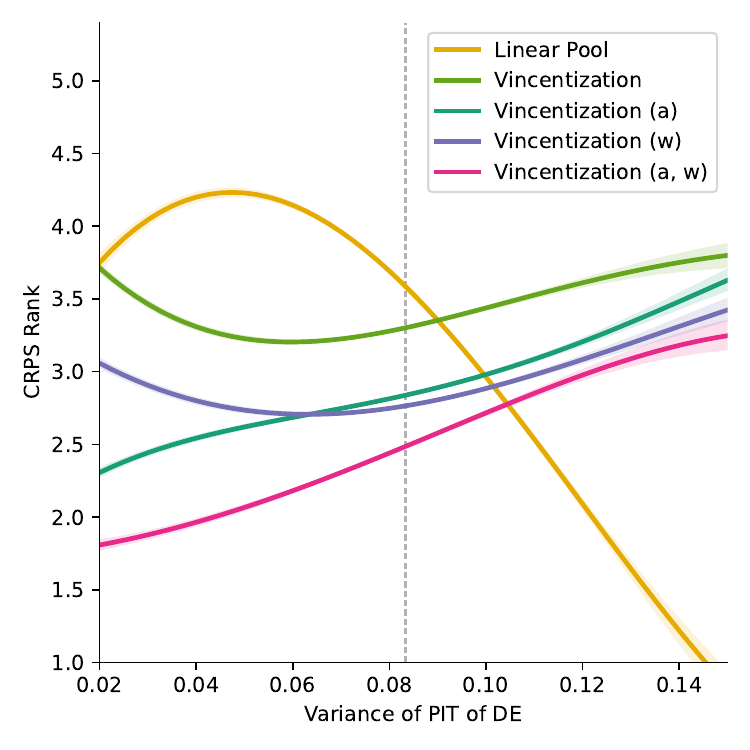}
		\caption{Polynomial regression curves of order 4 showing the relationship between the CRPS ranking of the aggregation methods and the PI coverage (left) resp.\ the dispersion (right) of the DE. 
        % The bands around the curves display 95\% bootstrap confidence intervals. 
        \label{fig:bench_cov_rank_analysis}}
\end{center}
\end{figure}

Now, we investigate how the properties of the DE forecast affect the ranking of the methods. For this, we analyzed the dependence of the CRPS ranking of the aggregation methods depending on the PI coverage of the DE. The left panel of Figure \ref{fig:bench_cov_rank_analysis} shows a curve for each aggregation method based on a polynomial regression analysis, which was carried out on the evaluation data and models the relationship between the CRPS ranking and the PI coverage of the DE. The curves show that the LP performs better than the VI variants when the PI coverages is below the nominal level but becomes worse as the coverage increases. These findings agree with the theoretical properties of the LP and the in-depth analysis, as underdispersed forecasts result in low PI coverages. For the VI variants, we observe the contrary effect. 

However, we consider only one nominal level for the PI coverage in this study. Therefore, we additionally analyze the performance in terms of the dispersion, which we define as the variance of the PIT in Section \ref{ssec:evaluation}. The right panel in Figure \ref{fig:bench_cov_rank_analysis} shows the results of an analogous regression analysis but based on the dispersion instead of the PI coverage. Recall that values below 0.0833 correspond to overdispersion and values above to underdispersion. The polynomial curves obtained by the regression analysis confirm the previous findings in that the LP works especially well for underdispersed DE forecasts. Again, we observe the contrary effect for the VI variants. Notably, for calibrated DE forecasts, the analysis indicates that the VI variants result in a better performance.

\begin{figure}
	\begin{center}
		\includegraphics[width=0.49\textwidth]{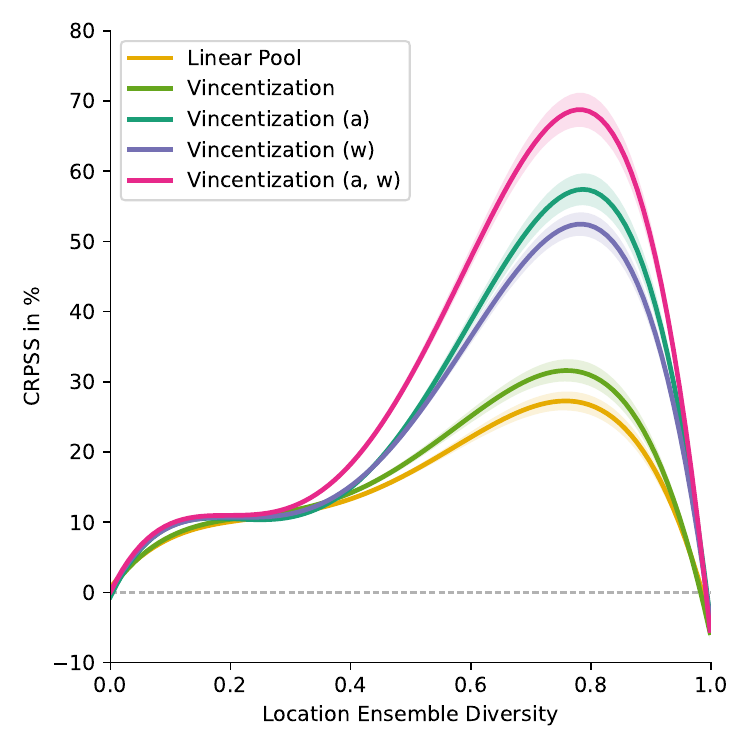}
		\caption{Polynomial regression curves of order 4 showing the relationship between the CRPSS of the aggregation methods and the location diversity of the DE. 
        \label{fig:bench_div_loc_analysis}}
\end{center}
\end{figure}

Next to the dispersion of the DE forecasts, the diversity within the DE may also be a relevant factor for the performance of the methods. 
While we did not find a connection of the ensemble diversity to the ranking of the aggregation methods, we did for the connection to the CRPSS.
Figure \ref{fig:bench_div_loc_analysis} shows the effect of the location diversity based on a regression analysis on the evaluation data.
In general, the skill of the aggregation methods increases as the diversity increases. Still, there is large spread in the relationship between diversity and CRPSS as the wide distribution of the individual cases shows. Interestingly, in cases where the location diversity becomes larger than 0.4, we see an additional increase in skill. Hence, when the DE becomes more diverse, the improvement obtained by aggregation increases further. In these cases, the improvement by VI with parameter estimation is larger than that of the LP and \Vo. At last, when the ensemble diversity becomes "too" large, the improvement vanishes. However, the conclusions drawn for cases with large diversity are based on a relatively small number of outliers. For the diversity in terms of the prediction uncertainty measured by the PI length and in terms of the performance measured by the CRPS (see Figure \ref{fig:bench_div_lgt_crps_analysis} in Appendix \ref{supl:figures}), we come to similar conclusions. As the performance diversity increases, the CRPSS of the aggregation methods increases, while the differences between the methods become larger. Recall that the CRPSS is calculated with respect to the mean score of the DE, hence the CRPSS is a relative and not an absolute performance measure. For the prediction uncertainty diversity, only a small effect is visible. Altogether, we conclude that location and performance diversity result in more improvement obtained from aggregation.

\begin{figure}
	\begin{center}
		\includegraphics[width=\textwidth]{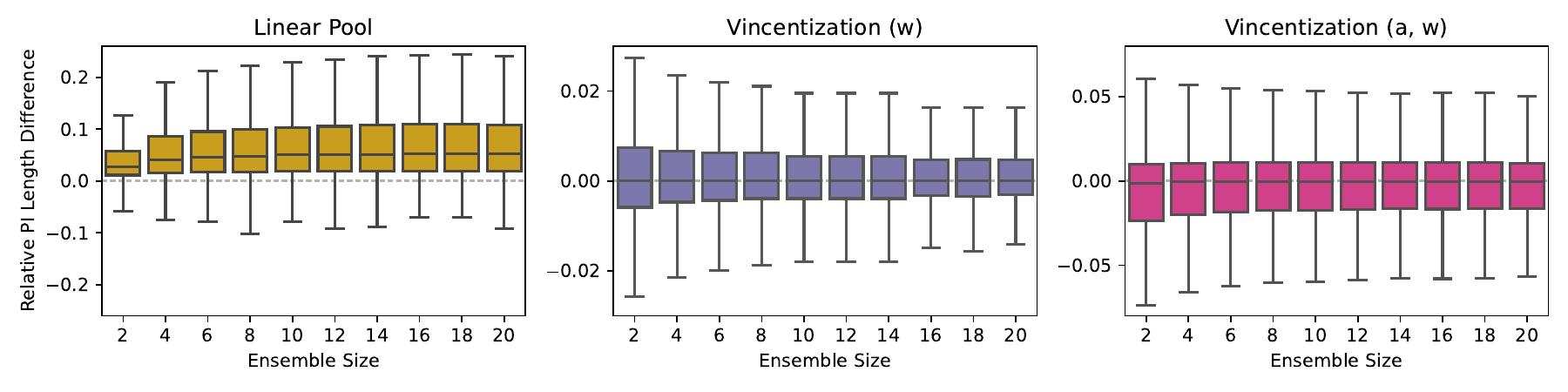}
		\caption{Boxplots of the relative PI length differences of the aggregation methods and the DE dependent on the ensemble size. Note the different scale of the y-axes. \label{fig:bench_n_ens_lgt}}
\end{center}
\end{figure}

We end the overall evaluation by analyzing the effect of the ensemble size on the performance of the aggregation methods. 
Figure \ref{fig:bench_n_ens_lgt} shows the relative PI length differences of the aggregation methods and the DE dependent on the ensemble size.
While the amplitudes of the relative PI length differences resemble those observed in Figure \ref{fig:bench_boxes_lgt_cov_nn}, we find that the PI length of the LP forecasts is more dependent on the ensemble size than that of the other two variants. For ensembles of size 2 to 10, the PI length increase with the size of the ensemble. A similar effect was observed for both data sets in the in-depth analysis. For \Vaw, we find that the PI length also increases, but to a smaller extent. However, the spread of the relative PI length differences decreases slightly for the two VI variants that include parameter estimation. For LP, this is not the case. As in the in-depth analysis, most effects are observed up to ensembles of size 10.

\begin{figure}
	\begin{center}
		\includegraphics[width=\textwidth]{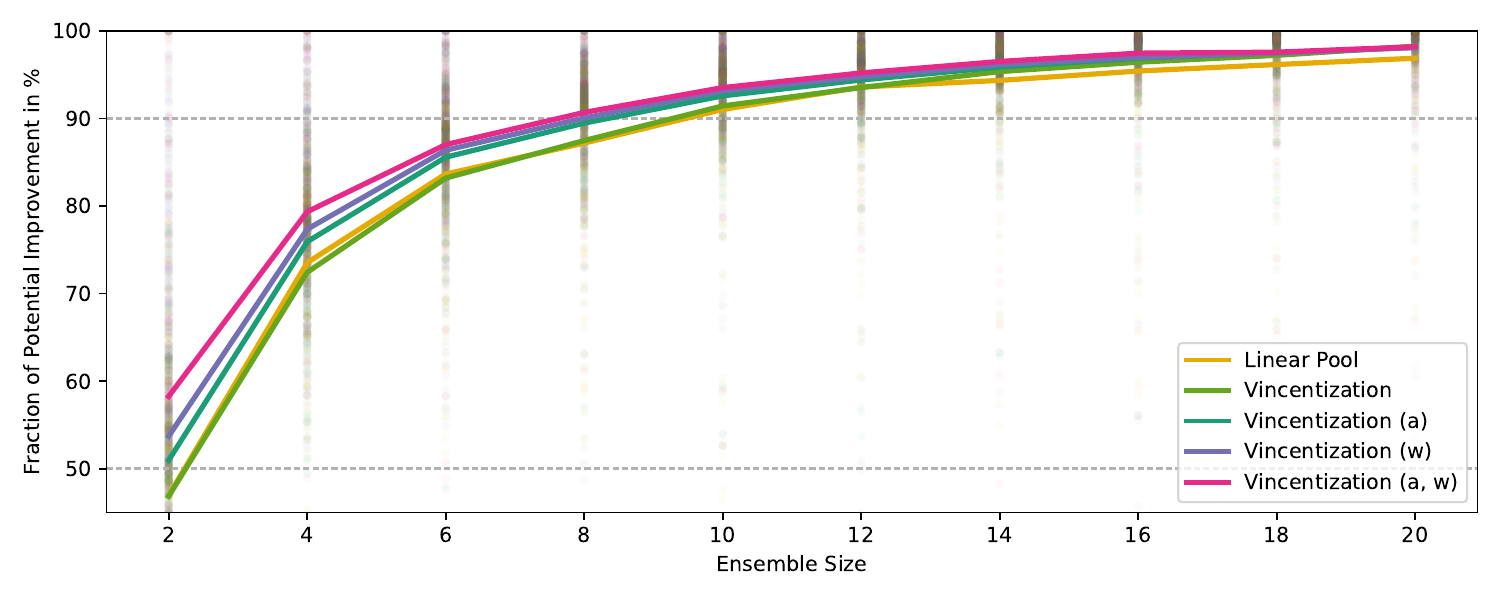}
		\caption{Relationship between the fraction of the potential improvement of the aggregation methods and the ensemble size. The curve is based on the mean values of the individual fractions that are derived for each data set, ensembling strategy and NN variant in addition to the aggregation method and ensemble size.\label{fig:bench_n_ens_skill}}
\end{center}
\end{figure}

The in-depth analysis showed that the improvement obtained by aggregation is saturated for ensembles of size 20. Here, we investigate whether this also holds for all benchmark data sets in general. 
For this, we compute the fraction of the potential improvement from the aggregation method that is already reached for the given ensemble size. The potential improvement is defined as the difference of the best CRPS value from all partitions and ensemble sizes of the corresponding setting with the CRPS value of the DE. For each ensemble size, we calculate the corresponding difference and set this in relation to the maximum improvement to compute the desired fraction.
Figure \ref{fig:bench_n_ens_skill} shows the fraction of the potential improvement dependent on the ensemble size. 
The plot reveals that an ensemble of size 2 improves upon the DE forecast by almost 50\% with respect to the potential improvement for \Vo\ and LP. For the VI variants with parameter estimation, the fraction is larger and almost 60\% for \Vaw. 
A reason why the fraction is larger for the VI variants with parameter estimation is that the aggregated forecasts improve upon the DE additionally due to the corrections applied in the generalized VI framework.
The improvement drastically increases by around 40\% to ensembles of size 8, where we already have around 90\% of the maximum possible improvement. Hence, by using an ensemble of size 8, one has already reached 90\% of the potential improvement from aggregation. Afterwards, the improvement increases by around 1--2\% for each step of 2 in the ensemble size. 
We did not observe any systematic differences in the dependence on the ensemble size for the NN variants and ensembling strategies.

\section{Discussion and conclusions} \label{sec:conclusion}

We have conducted a systematic comparison of aggregation methods for the combination of distribution forecasts from ensembles of neural networks based on different ensembling strategies, so-called deep ensembles.
In doing so, our work aims to reconcile and consolidate findings from the statistical literature on forecast combination and the machine learning literature on ensemble methods.
Specifically, we propose a general Vincentization framework where quantile functions of the forecast distributions can be flexibly combined, and compare to the results of the widely used linear pool, where the probabilistic forecasts are linearly combined on the scale of probabilities.
For deep ensembles of three variants of NN-based models for probabilistic forecasting that differ in the characterization of the output distribution, aggregation with both the LP and VI improves the predictive performance in a comprehensive evaluation on twelve data sets using seven ensembling strategies. 
The VI approaches frequently outperform the LP, but their ranking depends on the characteristics of the deep ensemble forecasts, especially the dispersion of the forecasts.
For example, given ensemble members that are already calibrated or overdispersed, the VI approaches are superior to the LP. While all approaches improve the predictive accuracy, the LP increases the dispersion of the forecasts resulting in (more) overdispersed forecasts.
If the individual forecast distributions are subject to systematic errors such as biases and dispersion errors, coefficient estimation via \Va, \Vw\ and \Vaw\ is able to correct these errors and improve the predictive performance considerably.
% , otherwise \Vo\ should be preferred. 
While these combination approaches require the estimation of additional combination coefficients, the computational costs are negligible compared to the generation of the NN-based probabilistic forecasts and can be performed on the validation data without restricting the estimation of the NNs. 
However, the smaller the validation data set, the larger the variability in the actual improvement from aggregation.
In terms of the ensembling strategies, we found that VI performs better than the LP for deep ensembles generated with one base-model, e.g., dropout or Bayesian NNs. In particular, VI with parameter estimation performs especially well for such deep ensembles.

Even though forecast combination generally improves the predictive performance, a lack of calibration of severely misspecified individual forecast distributions cannot be corrected by the aggregation methods considered here.
In the context of NNs and deep ensembles, the calibration of (ensemble) predictions and re-calibration procedures have been a focus of much recent research interest \citep{Guo2017,Ovadia2019}. 
For example, in line with the results of \cite{Gneiting2013}, deep ensemble predictions based on the LP were found to be miscalibrated and should be re-calibrated after the aggregation step \citep{Rahaman2020,Wu2021}.
A wide range of re-calibration methods, which simultaneously aggregate and calibrate the ensemble predictions (such as the \Va, \Vw\ and \Vaw\ approaches presented in Section \ref{sssec:agg_vincentization} for VI), have been proposed in order to correct the systematic errors introduced by the LP in the context of probability forecasting for binary events \citep{Allard2012}. For example, the beta-transformed LP composites the CDF of a Beta distribution with the LP \citep{Ranjan2010}, and \citet{Satopaa2014} propose to aggregate probabilities on a log-odds scale. Some of these approaches can be readily extended to the case of forecast distributions considered here \citep{Gneiting2013}.
For VI, more sophisticated approaches that allow the weights to depend on the quantile levels might improve the predictive performance \citep{Fakoor2023}. Further, moving from a linear combination function towards more complex transformations allowing for non-linearity might help to correct more involved calibration errors.

The focus on our study was on the effects of aggregating distribution forecasts from a given deep ensemble, and not on finding the overall best ensembling strategy to produce neural network-based probabilistic forecasts in the form of a deep ensemble.
While we did not systematically compare the performance of the ensembling strategies, the naive ensemble generally seemed to perform best in terms of the CRPS. In particular, the naive ensemble seemed to be the most stable approach for generating a deep ensemble. 
That said, the other ensembling strategies have distinct advantages and have proven their effectiveness in other applications
Most importantly, aggregating forecasts did not have an effect on the ranking of the ensembling strategies.

When deciding how to generate a deep ensemble, the trade-off between computational costs and predictive performance plays an important role. Larger ensembles yield a better predictive performance but at the expense of increased computing time. In case of base-model approaches, the additional computational costs are in general significantly lower compared to the multi-model approaches, where multiple models need to be trained. 
However, the uncertainty in the training of the base model is not taken account for.
To enhance predictive performance, one could follow both approaches by generating multiple base-models, which are then each used to generate their own sub-ensemble that need to aggregated altogether. 
While we have assumed equally weighted aggregation schemes, more sophisticated approaches that take the interplay of two ensemble-generating mechanisms into account might enhance the predictive performance. 

Finally, we summarize four key recommendations for aggregating distribution forecasts from deep ensembles based on our results. 
First, in order to optimize the final predictive performance of the aggregated forecast, the individual component forecasts should be optimized as much as possible.\footnote{
\cite{Abe2022} find that deep ensembles do not offer benefits compared to single larger (that is, more complex) NNs. Our results do not contradict their findings since we address a conceptually different question and argue that given the generation of a deep ensemble, the individual members' forecasts should be optimized as much as possible. In this situation, a single NN will generally not be able to match the predictive performance of the associated deep ensemble.
}
While forecast combination improves predictive performance, it generally did not effect the ranking of the different NN-variants for generating probabilistic forecasts, and can be unable to fix substantial systematic errors. 
Second, generating an ensemble with a size of 10 appears to be a sensible choice, with only minor improvements being observed for up to 20 members. 
This corresponds to the results in \cite{Fort2019} and ensemble sizes typically chosen in the literature \citep{Lakshminarayanan2017,Rasp2018}, but the benefits of generating more ensemble members need to be balanced against the computational costs, and sometimes smaller ensembles have been suggested \citep{Ovadia2019,Abe2022}.
Third, the choice of aggregation methods should take the dispersion of the individual ensemble member forecasts into account. For calibrated and overdispersed forecasts, VI is favorable, for underdispersed forecasts, the LP may be the better option.
Fourth and last, parameter estimation via \Vaw\ frequently enhances predictive performance, especially for larger data sets or base-model ensembling strategies.
The choice of the specific variant within the general VI framework depends on potential misspecifications of the individual component distributions.
Note that these conclusions, in particular the superiority of the quantile aggregation approaches, refer to the specific situation of deep ensembles considered here.
The property of shape-preservation justifies the use of VI from a theoretical perspective in a setting where the ensemble members are based on the same model and data.
If the ensemble members differ in terms of the model used to generate the forecast distribution or the input data they are based on, shape-preservation might not be desired. 
Instead, a model selection approach based on the LP, which allows for obtaining a multi-modal forecast distribution, might better represent the possible scenarios that may materialize.

\section*{Acknowledgments}

The research leading to these results has been done within the project C5 ``Dynamical feature-based ensemble postprocessing of wind gusts within European winter storms'' of the Transregional Collaborative Research Center SFB/TRR 165 ``Waves to Weather'' funded by the German Research Foundation (DFG). Sebastian Lerch gratefully acknowledges support by the Vector Stiftung through the Young Investigator Group ``Artificial Intelligence for Probabilistic Weather Forecasting''. We thank Daniel Wolffram, Eva-Maria Walz, Nina Horat, Anja Mühlemann, Alexander Jordan and Tilmann Gneiting for helpful comments and discussions.

\bibliographystyle{myims2}
\bibliography{paper_cpn}

\clearpage
\newpage

%
% start of Appendix
%

\setcounter{figure}{0}

\makeatletter 
\renewcommand{\thefigure}{S\@arabic\c@figure}
\makeatother

\setcounter{table}{0}

\makeatletter 
\renewcommand{\thetable}{S\@arabic\c@table}
\makeatother

\setcounter{section}{0}
\renewcommand{\thesection}{S\arabic{section}}

\section*{Supplementary material}

\section{Network setup} \label{supl:nn_setup}

Here, we describe the setup of the NNs in more detail. First, note that the predictor variables are standardized based on the respective training data (i.e., separately for each partition) in a preprocessing step. Other than that, all NNs are trained over 500 epochs using early stopping with a patience of 10. The NNs are implemented in Python \citep[3.10.6;][]{Python2022} via keras \citep[2.10.0;][]{keras2015} built on tensorflow \citep[2.10.0;][]{tensorflow2015}.

In case of the BatchEnsemble, we generate one ensemble of the maximum ensemble size, i.e., 20, and then use the first $n$ members for aggregation of an ensemble of size $n$ for each combination of NN variant, data set and partition. Further, the chosen batch sizes (see Table \ref{tbl:hpars}) refer to the effective batch sizes per ensemble member within the parallel training. If the required batch size for parallel training exceeds the size of the training set, we resample the missing data points.
In case of the dropout variants and BNN, where the ensemble is generated on one base model, the direct NN output of one prediction corresponds to one ensemble member. 
For the dropout variants, we drop only the neurons in the hidden layers and not the input layer. In case of MC dropout, the chosen architectures (see Table \ref{tbl:hpars}) refer to the effective architecture, as we additionally scale up the number of neurons based on the dropout rate.

Regarding the NN variants, the BQN models use 99 equidistant quantile levels from 0.01 to 0.99 in the loss function. For HEN, the target variable is also standardized on the training data (analogous to the predictor variables). As described in Section \ref{ssec:nn_hen}, the bin edges are defined by quantiles of a standard normal distribution. Based on experiments on the validation set and previous applications, 
we use equidistant quantile levels within the interval $\left[ 0.05, 0.95 \right]$ and a finer resolution (with respect to the quantile level) in the tails of the distribution. For the tails, we chose the 10 (fixed) bin edges $b_0 = \Phi^{-1} \left( 10^{-16} \right)$, $b_1 = \Phi^{-1} \left( 10^{-8} \right)$, $b_2 = \Phi^{-1} \left( 0.0001 \right)$, $b_3 = \Phi^{-1} \left( 0.01 \right)$, $b_4 = \Phi^{-1} \left( 0.05 \right)$, where $\Phi$ denotes the CDF of the standard normal distribution, and $b_{N+1-\ell} = 1-b_{\ell}$ for $\ell = 0, \dots, 4$.
If the minimum (maximum) within the training data is smaller (larger) than the bin edge $b_0$ ($b_{N+1}$), we adapt the bin edge to this value minus (plus) a small threshold. The other $N - 9$ bin edges are then chosen as the quantiles at equidistant levels between 0.05 and 0.95.

\section{Hyperparameter tuning} \label{supl:hpar_tuning}

For the hyperparameter tuning, we first note that we did not perform a separate hyperparameter tuning for bagging and BatchEnsemble, but instead use the hyperparameters obtained for the naive ensemble runs. This was done, as we tune the performance of an individual ensemble member, which are structurally identical for these three variants. 
% Note that the tuned batch size corresponds to the effective batch size for the BatchEnsemble.
As described in Section \ref{sec:cs}, we choose the hyperparameters that perform best on the first two random partitions. Performance was measured based on the mean CRPS and sanity checked with PIT histograms and the logarithmic score (negative log-likelihood). Unless a severe degree of miscalibration or strong deviations in the logarithmic score were detected (with respect to the competing hyperparameter sets), the hyperparameters with the lowest CRPS were chosen. The following variables and values were considered for hyperparameter tuning:
\begin{itemize}
    \item Batch size (BA): 16, 32, 64, 256.
    \item Activation function (Actv): Relu, Softplus.
    % \item Architectures (Arch): [64, 32] (denoted by 2--64 in Tables \ref{tbl:hpars0} and \ref{tbl:hpars1}), [512, 256] (2--512), [64, 64, 32] (3--64), [512, 512, 256] (3--512), [512, 512, 256, 128] (4--512).
    \item Architectures (Arch): [64, 32], [512, 256], [64, 64, 32], [512, 512, 256], [512, 512, 256, 128]. In Table \ref{tbl:hpars}, we denote the architecture by the number of layers and the number of nodes in the first layer, e.g., 2--512 for [512, 256].
    \item Learning rate (LR): 0.001, 0.0005.
    \item Dropout rate (DR; for MC dropout): 5\%, 10\%, 20\%, 50\%, 80\%.
    \item Prior (PR; for Bayesian NN): Uniform, standard normal, Laplace.
    \item Degree of Bernstein polynomials $d$ (for BQN): 8, 12.
    \item Number of bins $N$ (for HEN): 20, 30.
\end{itemize}
Tables \ref{tbl:hpars} lists the chosen hyperparameter configurations. Recall that all NNs are trained over 500 epochs using early stopping with a patience of 10.

\begin{table}
\caption{
Hyperparameter choices for the case studies in Section \ref{sec:cs}. For the notation, see Appendix \ref{supl:hpar_tuning}. 
% Arch denotes the architecture chosen among the variants presented in Appendix \ref{supl:nn_hpar}, Actv the activation function, BA the batch size, LR the learning rate, $d$ the degree of the Bernstein polynomials for BQN and $N$ the number of bins for HEN. 
\label{tbl:hpars}}	
\begin{center}
\scalebox{0.62}{\input{hpars.txt}}
\end{center}
\end{table}

\section{Description of the simulation studies} \label{supl:sim_studies}

Scenarios 1 and 2 in Section \ref{ssec:cs_bench} correspond to models 1 and 4 proposed in \citet{Li2021}.
The results for their other models do not provide additional insights and are thus not included here. Note that we reduce the size of the test set from 10,000 to 1,000 for computational reasons.

The first simulation scenario we consider is a linear model with normally distributed errors. Based on a random vector of predictors $\boldsymbol{X} \in \R^5$, which serves as the input of the NNs, and the random coefficient vectors $\boldsymbol{\beta_1}, \boldsymbol{\beta_2} \in \R^5$, which are fixed for each run of the simulation and unknown to the forecaster, the target variable $Y$ is calculated via
\begin{eqnarray}
Y = \boldsymbol{X}^T \boldsymbol{\beta_1} + \epsilon \cdot \exp \left( \boldsymbol{X}^T \boldsymbol{\beta_2} \right), \nonumber 
\end{eqnarray}
where $\boldsymbol{X} \sim \mathcal{N} \left( \boldsymbol{0}, \boldsymbol{I_5} \right)$, $\boldsymbol{\beta_1} \sim \mathcal{N} \left(\boldsymbol{0}, \boldsymbol{I_5}\right)$, $\boldsymbol{\beta_2} \sim \mathcal{N} \left(\boldsymbol{0}, 0.45^2\boldsymbol{I_5}\right)$
and $\epsilon \sim \mathcal{N} \left(0, 1\right)$. 
% The optimal forecast $F^*$ is then given by 
% \begin{eqnarray}
% 	Y \mid \boldsymbol{X}, \boldsymbol{\beta_1}, \boldsymbol{\beta_2} \; \sim \; \mathcal{N} \left( \boldsymbol{X}^T \boldsymbol{\beta_1}, \exp \left( 2\boldsymbol{X}^T \boldsymbol{\beta_2} \right) \right) = F^*. \nonumber 
% \end{eqnarray}
In the second scenario, we consider a skewed distribution with a nonlinear mean function.
The target variable $Y$ is defined by
\begin{eqnarray}
	Y = 10 \sin \left( 2\pi X_1 X_2 \right) + 20 \left( X_3 - 0.5 \right)^2 + 10 X_4 + 5 X_5 + \epsilon, \nonumber 
\end{eqnarray}
where $\boldsymbol{X} = \left( X_1, \dots, X_5 \right)^T$, $X_1, \dots, X_5 \overset{iid}{\sim} \mathcal{U} \left(0, 1\right)$, and $\epsilon \sim \text{SkewNormal} \left(0, 1, -5\right)$. 
% The optimal forecast $F^*$ is given by 
% \begin{eqnarray}
% 	Y \mid \boldsymbol{X} \; \sim \; \text{SkewNormal} \left( 10 \sin \left( 2\pi X_1 X_2 \right) + 20 \left( X_3 - 0.5 \right)^2 + 10 X_4 + 5 X_5, 1, -5 \right) = F^*. \nonumber 
% \end{eqnarray}

\clearpage
\section{Additional figures} \label{supl:figures}

\begin{figure}[!h]
	\begin{center}
		\includegraphics[width=\textwidth]{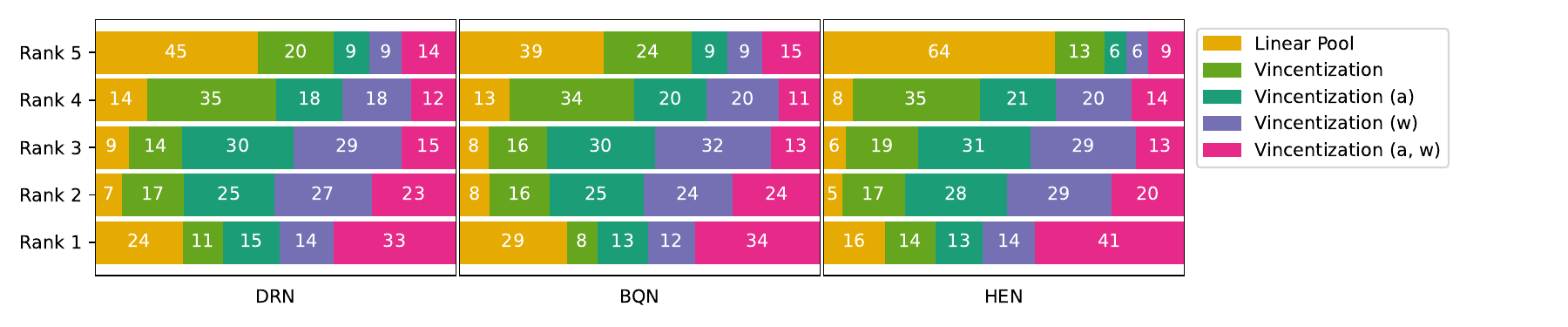}
		\includegraphics[width=\textwidth]{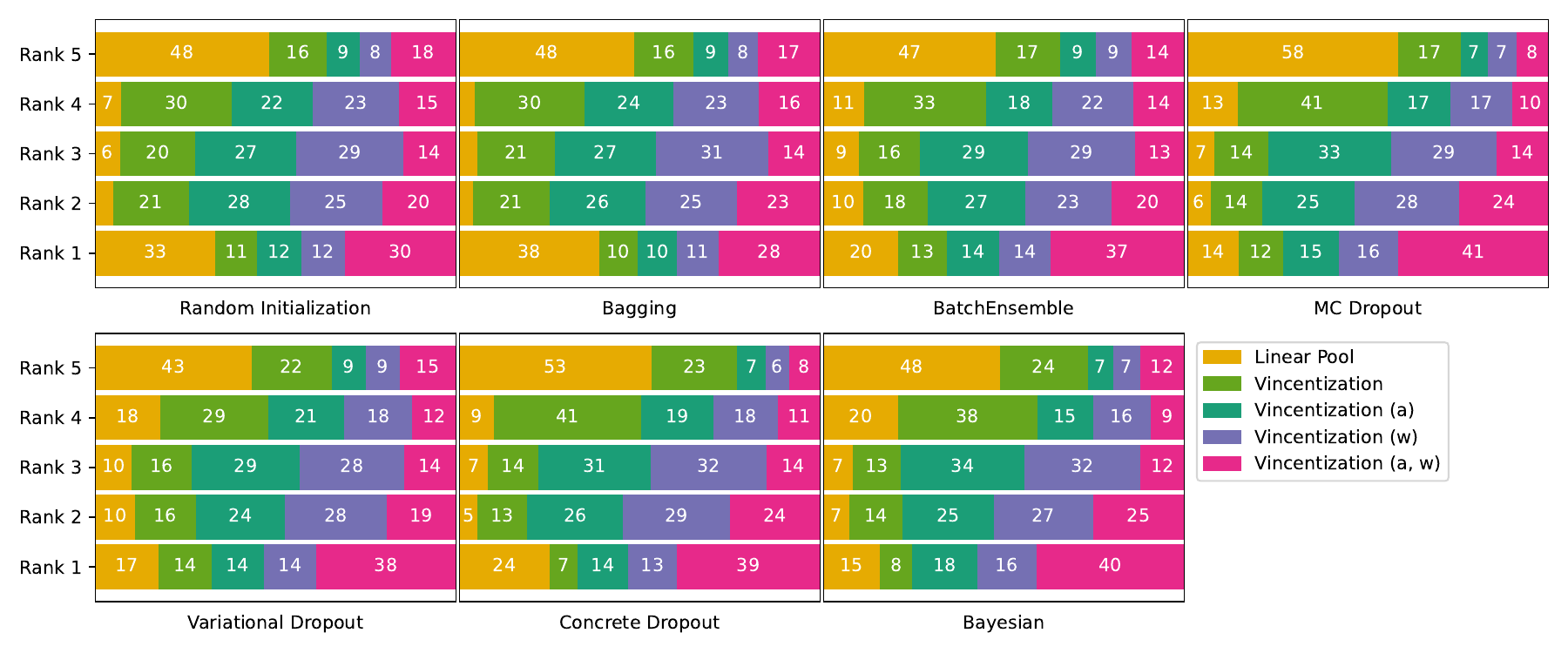}
		\includegraphics[width=\textwidth]{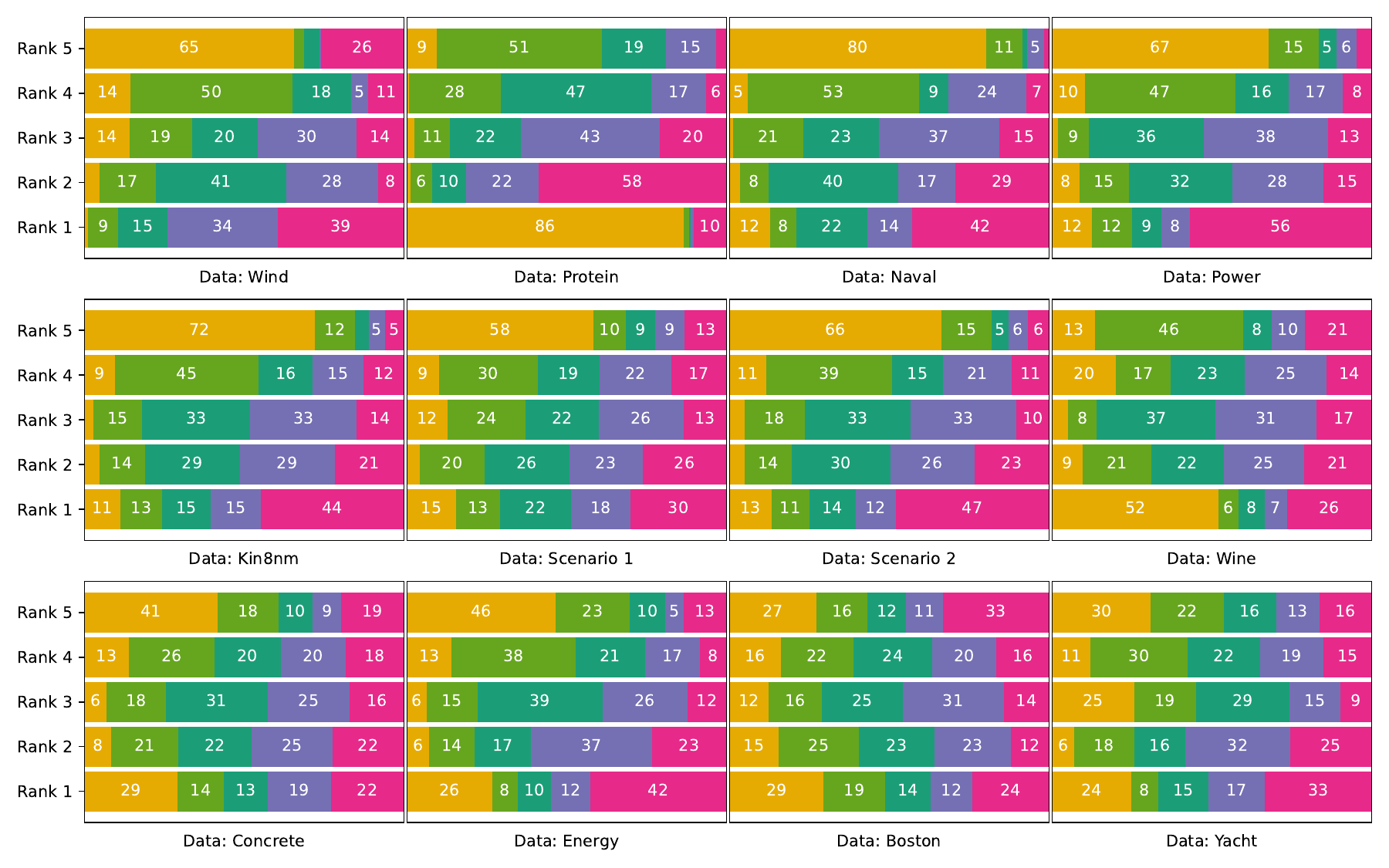}
		\caption{Stacked bar plots showing the relative distribution of the CRPS ranking dependent on the NN variant, ensembling strategy and data set. Percentages below 5\% are not labeled. The data sets are ordered according to their size starting with the largest. \label{fig:bench_rank_barplots}}
    \end{center}
\end{figure}

\begin{figure}
	\begin{center}
		\includegraphics[width=\textwidth]{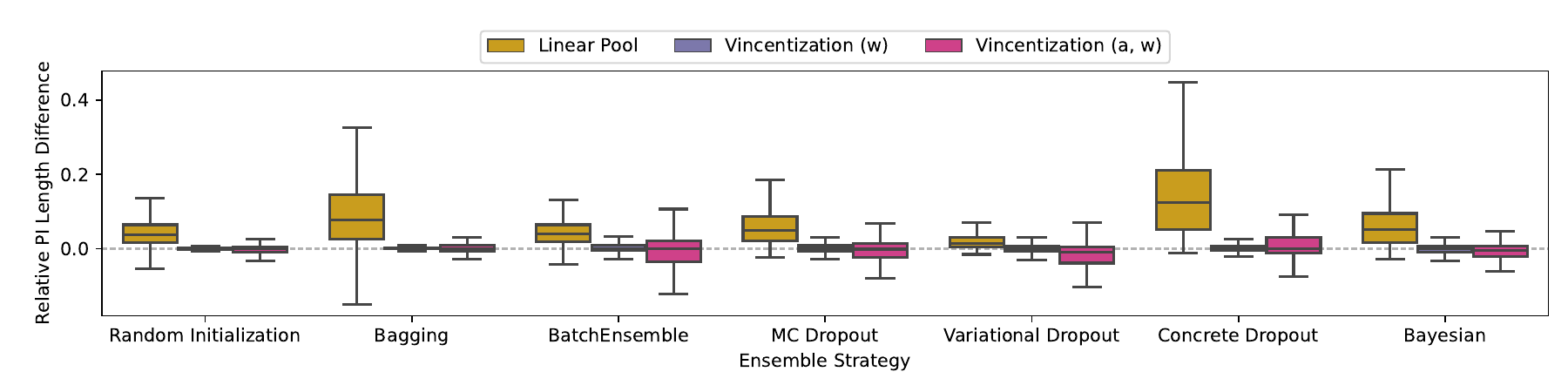}
		\includegraphics[width=\textwidth]{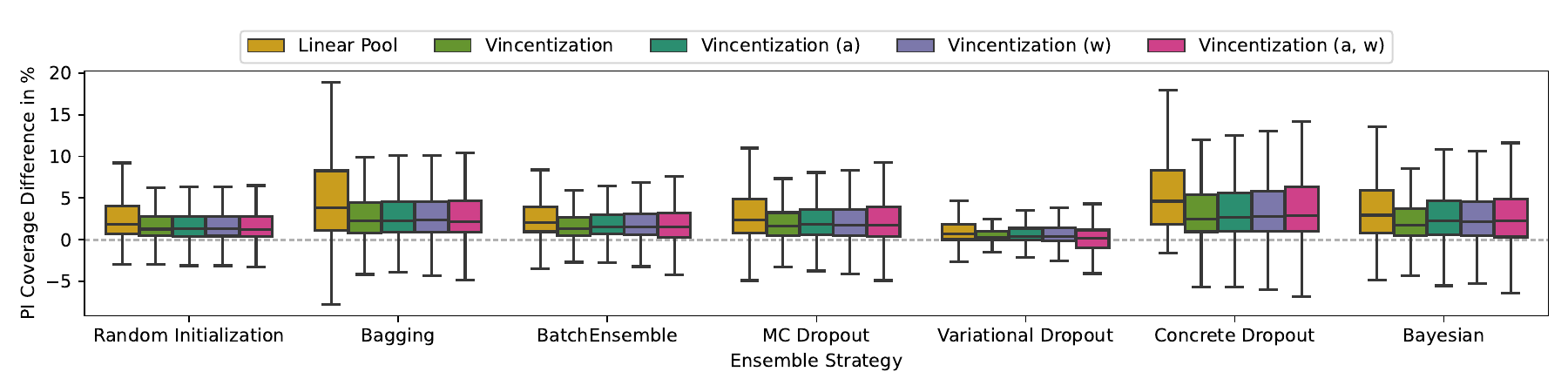}
		\caption{Boxplots of the relative PI length differences (left) and the PI coverage differences (right) with respect to the DE dependent on the aggregation method and ensembling strategy. \label{fig:bench_boxes_lgt_cov_ens}}
\end{center}
\end{figure}

\begin{figure}
	\begin{center}
		\includegraphics[width=\textwidth]{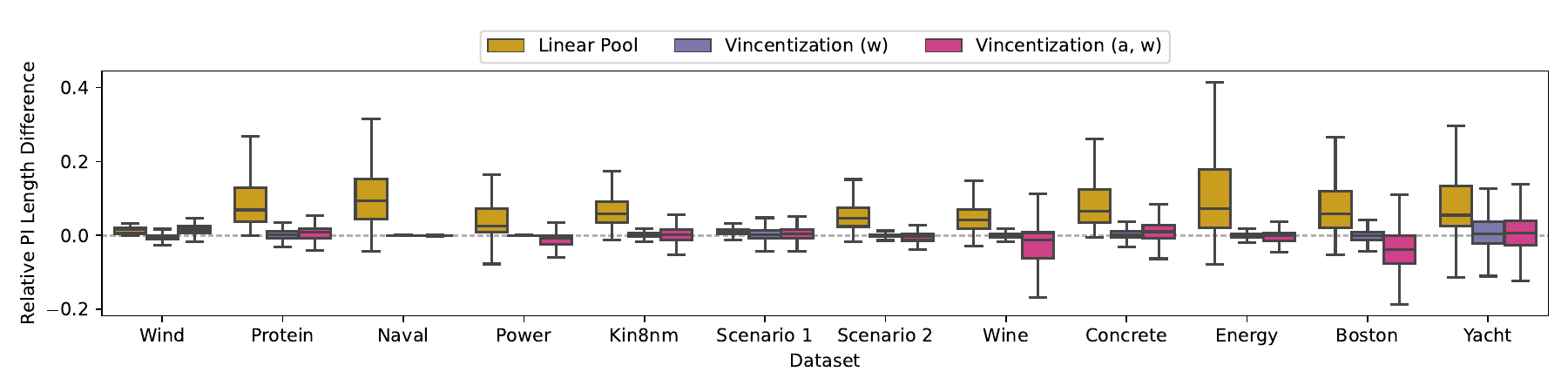}
		\includegraphics[width=\textwidth]{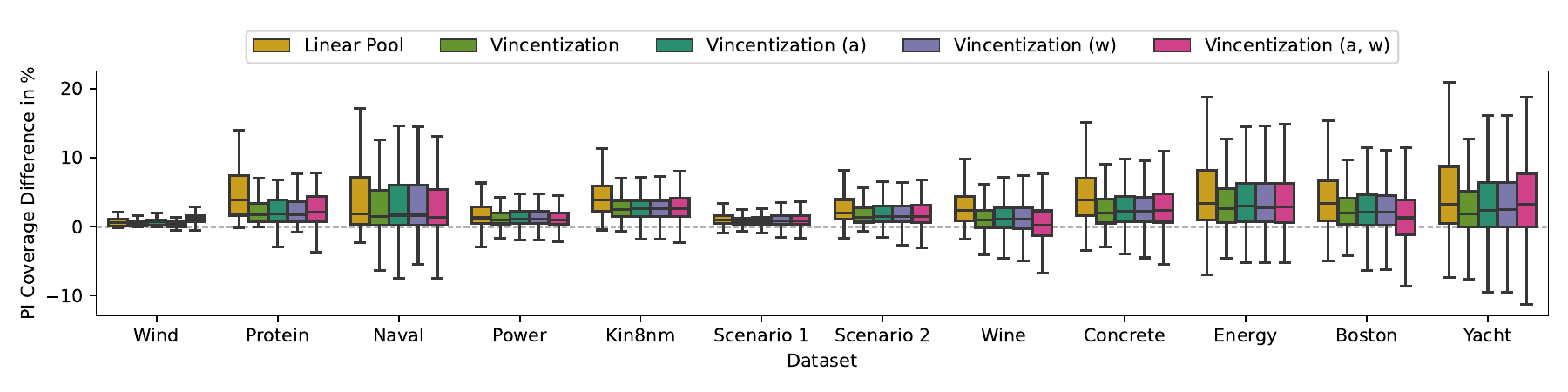}
		\caption{Boxplots of the relative PI length differences (left) and the PI coverage differences (right) with respect to the DE dependent on the aggregation method and data set. \label{fig:bench_boxes_lgt_cov_datasets}}
\end{center}
\end{figure}

\begin{figure}
	\begin{center}
		\includegraphics[width=0.49\textwidth]{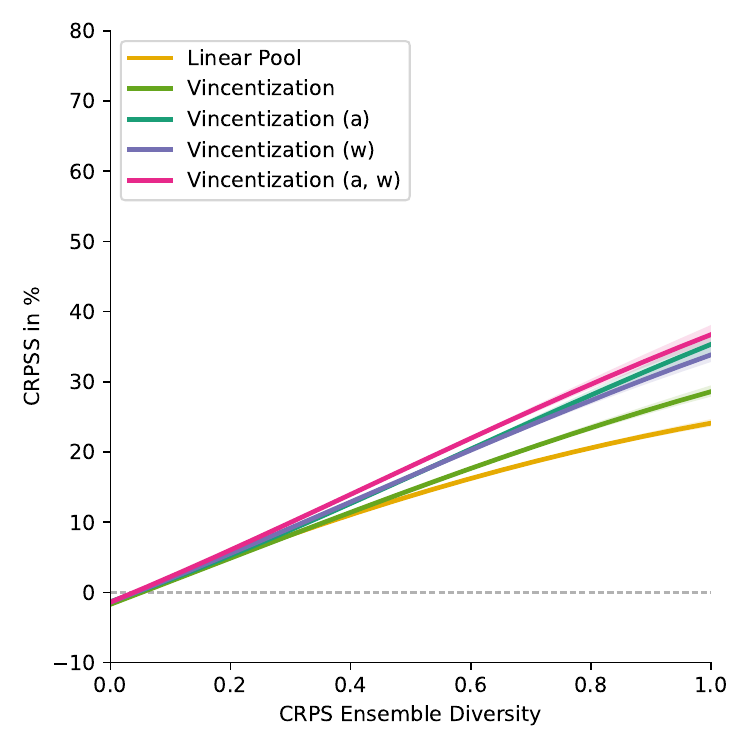}
		\includegraphics[width=0.49\textwidth]{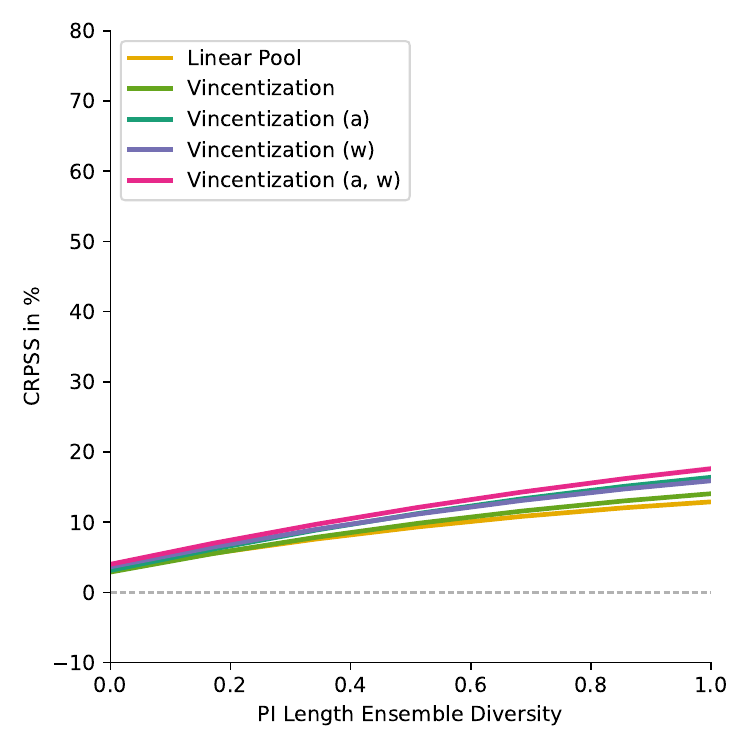}
		\caption{Polynomial regression curves of order 4 showing the relationship between the CRPSS and the prediction performance (left) resp.\ uncertainty (right) diversity of the aggregation methods. 
        % The bands around the curves display 95\% bootstrap confidence intervals.
        \label{fig:bench_div_lgt_crps_analysis}}
\end{center}
\end{figure}

\end{document}

%% file: hpars.txt
\begin{tabular}{l*{7}{r}|*{7}{r}|*{6}{r}} 
\toprule 
&& \multicolumn{5}{c}{DRN} & \multicolumn{1}{c}{} & \multicolumn{6}{c}{BQN} & \multicolumn{1}{c}{} & \multicolumn{6}{c}{HEN} \\  
\cmidrule{3-7} \cmidrule{9-14} \cmidrule{16-21}  
&& Arch & Actv & BA & LR & DR/PR & \multicolumn{1}{c}{} & $d$ & Arch & Actv & BA & LR & DR/PR & \multicolumn{1}{c}{} & $N$ & Arch & Actv & BA & LR & DR/PR \\ 
\midrule 
\multicolumn{7}{l}{\texttt{Naive Ensemble / Bagging / BatchEnsemble}} & & & & & & & & & & & & \\ 
Gusts 
 & & 2--512 & Soft & 32 & .0010 & - 
 & & 12  & 3--512 & Soft & 64 & .0010 & - 
 & & 20  & 4--512 & Soft & 64 & .0005 & - 
 \\ 
Scenario 1 
 & & 2--512 & Soft & 16 & .0005 & - 
 & & 8  & 4--512 & Soft & 64 & .0010 & - 
 & & 30  & 4--512 & Soft & 16 & .0005 & - 
 \\ 
Scenario 2 
 & & 2--512 & Soft & 256 & .0005 & - 
 & & 8  & 3--512 & Soft & 16 & .0005 & - 
 & & 30  & 4--512 & Soft & 64 & .0005 & - 
 \\ 
Protein 
 & & 4--512 & Relu & 16 & .0005 & - 
 & & 8  & 4--512 & Relu & 32 & .0010 & - 
 & & 30  & 4--512 & Relu & 32 & .0010 & - 
 \\ 
Naval 
 & & 4--512 & Relu & 16 & .0010 & - 
 & & 8  & 2--512 & Relu & 16 & .0005 & - 
 & & 30  & 3--\phantom{{0}}64 & Relu & 16 & .0005 & - 
 \\ 
Power 
 & & 3--\phantom{{0}}64 & Relu & 16 & .0005 & - 
 & & 12  & 3--\phantom{{0}}64 & Relu & 256 & .0005 & - 
 & & 30  & 4--512 & Relu & 16 & .0005 & - 
 \\ 
Kin8nm 
 & & 4--512 & Soft & 16 & .0005 & - 
 & & 8  & 3--\phantom{{0}}64 & Soft & 16 & .0005 & - 
 & & 20  & 3--\phantom{{0}}64 & Soft & 16 & .0010 & - 
 \\ 
Wine 
 & & 2--512 & Relu & 16 & .0010 & - 
 & & 12  & 3--512 & Relu & 16 & .0010 & - 
 & & 30  & 3--\phantom{{0}}64 & Relu & 16 & .0005 & - 
 \\ 
Concrete 
 & & 3--512 & Relu & 16 & .0010 & - 
 & & 8  & 3--512 & Relu & 16 & .0005 & - 
 & & 20  & 4--512 & Relu & 16 & .0005 & - 
 \\ 
Energy 
 & & 2--512 & Relu & 16 & .0010 & - 
 & & 12  & 2--512 & Relu & 16 & .0005 & - 
 & & 30  & 4--512 & Soft & 16 & .0010 & - 
 \\ 
Boston 
 & & 3--512 & Relu & 16 & .0010 & - 
 & & 8  & 4--512 & Relu & 16 & .0005 & - 
 & & 30  & 4--512 & Relu & 16 & .0010 & - 
 \\ 
Yacht 
 & & 2--\phantom{{0}}64 & Soft & 16 & .0010 & - 
 & & 12  & 3--\phantom{{0}}64 & Soft & 16 & .0005 & - 
 & & 30  & 3--512 & Soft & 16 & .0010 & - 
 \\ 
\midrule 
\multicolumn{7}{l}{\texttt{MC Dropout}} & & & & & & & & & & & & \\ 
Gusts 
 & & 2--512 & Soft & 32 & .0005 & 20\% 
 & & 8  & 3--\phantom{{0}}64 & Soft & 16 & .0005 & 10\% 
 & & 20  & 2--\phantom{{0}}64 & Soft & 32 & .0005 & 10\% 
 \\ 
Scenario 1 
 & & 2--512 & Soft & 16 & .0010 & 5\% 
 & & 8  & 3--512 & Soft & 16 & .0010 & 5\% 
 & & 30  & 3--512 & Soft & 16 & .0010 & 5\% 
 \\ 
Scenario 2 
 & & 2--512 & Relu & 16 & .0010 & 5\% 
 & & 12  & 3--512 & Relu & 16 & .0005 & 5\% 
 & & 30  & 4--512 & Relu & 32 & .0005 & 5\% 
 \\ 
Protein 
 & & 4--512 & Relu & 32 & .0005 & 20\% 
 & & 8  & 4--512 & Relu & 64 & .0010 & 10\% 
 & & 30  & 4--512 & Relu & 64 & .0005 & 5\% 
 \\ 
Naval 
 & & 2--\phantom{{0}}64 & Relu & 16 & .0005 & 5\% 
 & & 8  & 4--512 & Relu & 16 & .0010 & 10\% 
 & & 30  & 3--512 & Relu & 16 & .0005 & 5\% 
 \\ 
Power 
 & & 2--512 & Soft & 16 & .0010 & 5\% 
 & & 8  & 2--512 & Soft & 16 & .0005 & 5\% 
 & & 30  & 3--512 & Relu & 32 & .0010 & 5\% 
 \\ 
Kin8nm 
 & & 3--512 & Soft & 32 & .0005 & 5\% 
 & & 8  & 4--512 & Soft & 16 & .0005 & 5\% 
 & & 30  & 4--512 & Soft & 16 & .0005 & 5\% 
 \\ 
Wine 
 & & 2--512 & Relu & 32 & .0010 & 5\% 
 & & 8  & 2--512 & Relu & 16 & .0005 & 5\% 
 & & 30  & 3--\phantom{{0}}64 & Relu & 64 & .0010 & 5\% 
 \\ 
Concrete 
 & & 4--512 & Relu & 16 & .0005 & 5\% 
 & & 12  & 3--512 & Relu & 16 & .0010 & 5\% 
 & & 20  & 4--512 & Relu & 16 & .0010 & 5\% 
 \\ 
Energy 
 & & 2--512 & Relu & 16 & .0010 & 5\% 
 & & 12  & 3--512 & Relu & 16 & .0005 & 5\% 
 & & 30  & 4--512 & Soft & 32 & .0010 & 10\% 
 \\ 
Boston 
 & & 4--512 & Relu & 16 & .0005 & 10\% 
 & & 12  & 3--512 & Relu & 32 & .0010 & 5\% 
 & & 30  & 4--512 & Relu & 16 & .0010 & 5\% 
 \\ 
Yacht 
 & & 2--512 & Relu & 16 & .0005 & 5\% 
 & & 12  & 2--512 & Relu & 16 & .0010 & 5\% 
 & & 30  & 3--512 & Soft & 16 & .0005 & 5\% 
 \\ 
\midrule 
\multicolumn{7}{l}{\texttt{Variational Dropout}} & & & & & & & & & & & & \\ 
Gusts 
 & & 2--\phantom{{0}}64 & Soft & 16 & .0010 & - 
 & & 12  & 2--\phantom{{0}}64 & Soft & 16 & .0005 & - 
 & & 20  & 3--\phantom{{0}}64 & Soft & 16 & .0010 & - 
 \\ 
Scenario 1 
 & & 2--\phantom{{0}}64 & Soft & 16 & .0010 & - 
 & & 8  & 2--\phantom{{0}}64 & Soft & 16 & .0005 & - 
 & & 30  & 2--\phantom{{0}}64 & Soft & 16 & .0010 & - 
 \\ 
Scenario 2 
 & & 3--\phantom{{0}}64 & Relu & 16 & .0010 & - 
 & & 12  & 2--\phantom{{0}}64 & Relu & 16 & .0010 & - 
 & & 30  & 3--512 & Soft & 16 & .0005 & - 
 \\ 
Protein 
 & & 2--\phantom{{0}}64 & Relu & 16 & .0010 & - 
 & & 12  & 3--512 & Relu & 32 & .0005 & - 
 & & 30  & 2--512 & Relu & 16 & .0005 & - 
 \\ 
Naval 
 & & 2--\phantom{{0}}64 & Soft & 64 & .0005 & - 
 & & 12  & 3--512 & Relu & 16 & .0010 & - 
 & & 20  & 3--512 & Soft & 16 & .0005 & - 
 \\ 
Power 
 & & 2--\phantom{{0}}64 & Relu & 32 & .0005 & - 
 & & 8  & 3--\phantom{{0}}64 & Soft & 16 & .0010 & - 
 & & 30  & 4--512 & Soft & 16 & .0005 & - 
 \\ 
Kin8nm 
 & & 2--\phantom{{0}}64 & Relu & 16 & .0010 & - 
 & & 12  & 2--\phantom{{0}}64 & Soft & 16 & .0005 & - 
 & & 30  & 3--\phantom{{0}}64 & Soft & 16 & .0010 & - 
 \\ 
Wine 
 & & 2--\phantom{{0}}64 & Relu & 16 & .0010 & - 
 & & 8  & 2--\phantom{{0}}64 & Soft & 16 & .0005 & - 
 & & 30  & 2--\phantom{{0}}64 & Relu & 16 & .0005 & - 
 \\ 
Concrete 
 & & 2--\phantom{{0}}64 & Soft & 16 & .0010 & - 
 & & 12  & 2--\phantom{{0}}64 & Relu & 16 & .0005 & - 
 & & 30  & 2--512 & Relu & 256 & .0005 & - 
 \\ 
Energy 
 & & 2--\phantom{{0}}64 & Relu & 16 & .0005 & - 
 & & 8  & 2--\phantom{{0}}64 & Relu & 16 & .0010 & - 
 & & 30  & 2--\phantom{{0}}64 & Relu & 32 & .0005 & - 
 \\ 
Boston 
 & & 2--\phantom{{0}}64 & Soft & 16 & .0005 & - 
 & & 12  & 2--\phantom{{0}}64 & Relu & 16 & .0010 & - 
 & & 30  & 2--\phantom{{0}}64 & Relu & 256 & .0010 & - 
 \\ 
Yacht 
 & & 2--\phantom{{0}}64 & Relu & 16 & .0010 & - 
 & & 12  & 2--512 & Relu & 32 & .0005 & - 
 & & 30  & 2--\phantom{{0}}64 & Relu & 16 & .0010 & - 
 \\ 
\midrule 
\multicolumn{7}{l}{\texttt{Concrete Dropout}} & & & & & & & & & & & & \\ 
Gusts 
 & & 3--\phantom{{0}}64 & Soft & 256 & .0005 & - 
 & & 8  & 3--512 & Soft & 32 & .0005 & - 
 & & 20  & 3--\phantom{{0}}64 & Soft & 16 & .0005 & - 
 \\ 
Scenario 1 
 & & 3--\phantom{{0}}64 & Soft & 16 & .0010 & - 
 & & 8  & 3--\phantom{{0}}64 & Soft & 16 & .0010 & - 
 & & 30  & 3--\phantom{{0}}64 & Soft & 32 & .0010 & - 
 \\ 
Scenario 2 
 & & 3--\phantom{{0}}64 & Soft & 16 & .0010 & - 
 & & 8  & 3--\phantom{{0}}64 & Soft & 16 & .0010 & - 
 & & 30  & 3--\phantom{{0}}64 & Relu & 16 & .0005 & - 
 \\ 
Protein 
 & & 4--512 & Relu & 32 & .0005 & - 
 & & 12  & 4--512 & Relu & 256 & .0010 & - 
 & & 30  & 4--512 & Relu & 64 & .0005 & - 
 \\ 
Naval 
 & & 2--512 & Relu & 256 & .0005 & - 
 & & 8  & 2--512 & Relu & 64 & .0005 & - 
 & & 30  & 3--\phantom{{0}}64 & Relu & 16 & .0005 & - 
 \\ 
Power 
 & & 3--512 & Relu & 16 & .0005 & - 
 & & 12  & 3--512 & Relu & 64 & .0010 & - 
 & & 30  & 4--512 & Relu & 16 & .0010 & - 
 \\ 
Kin8nm 
 & & 3--512 & Relu & 64 & .0005 & - 
 & & 12  & 3--512 & Relu & 32 & .0005 & - 
 & & 30  & 3--\phantom{{0}}64 & Soft & 16 & .0010 & - 
 \\ 
Wine 
 & & 3--512 & Relu & 32 & .0005 & - 
 & & 8  & 3--512 & Relu & 32 & .0005 & - 
 & & 30  & 4--512 & Relu & 32 & .0005 & - 
 \\ 
Concrete 
 & & 4--512 & Relu & 16 & .0010 & - 
 & & 8  & 3--512 & Relu & 32 & .0010 & - 
 & & 20  & 4--512 & Relu & 16 & .0005 & - 
 \\ 
Energy 
 & & 3--512 & Relu & 64 & .0010 & - 
 & & 8  & 2--512 & Relu & 16 & .0010 & - 
 & & 30  & 3--512 & Relu & 32 & .0010 & - 
 \\ 
Boston 
 & & 3--512 & Relu & 16 & .0010 & - 
 & & 12  & 3--512 & Relu & 64 & .0010 & - 
 & & 20  & 4--512 & Relu & 16 & .0005 & - 
 \\ 
Yacht 
 & & 4--512 & Relu & 16 & .0010 & - 
 & & 12  & 3--512 & Relu & 16 & .0010 & - 
 & & 20  & 3--512 & Soft & 32 & .0010 & - 
 \\ 
\midrule 
\multicolumn{7}{l}{\texttt{Bayesian}} & & & & & & & & & & & & \\ 
Gusts 
 & & 3--\phantom{{0}}64 & Soft & 32 & .0005 & Lapl 
 & & 12  & 3--512 & Soft & 64 & .0010 & Norm 
 & & 30  & 4--512 & Soft & 64 & .0010 & Lapl 
 \\ 
Scenario 1 
 & & 2--\phantom{{0}}64 & Soft & 32 & .0010 & Unif 
 & & 12  & 3--512 & Soft & 64 & .0010 & Norm 
 & & 30  & 3--512 & Soft & 64 & .0010 & Norm 
 \\ 
Scenario 2 
 & & 2--512 & Soft & 64 & .0005 & Unif 
 & & 12  & 2--512 & Soft & 16 & .0005 & Unif 
 & & 30  & 4--512 & Soft & 64 & .0005 & Norm 
 \\ 
Protein 
 & & 3--512 & Relu & 16 & .0010 & Norm 
 & & 8  & 4--512 & Relu & 32 & .0005 & Norm 
 & & 30  & 2--512 & Relu & 32 & .0005 & Norm 
 \\ 
Naval 
 & & 3--512 & Soft & 32 & .0010 & Unif 
 & & 8  & 3--512 & Soft & 32 & .0010 & Unif 
 & & 30  & 4--512 & Relu & 16 & .0005 & Norm 
 \\ 
Power 
 & & 2--\phantom{{0}}64 & Relu & 16 & .0005 & Unif 
 & & 8  & 3--\phantom{{0}}64 & Relu & 256 & .0005 & Lapl 
 & & 30  & 4--512 & Relu & 64 & .0010 & Norm 
 \\ 
Kin8nm 
 & & 4--512 & Relu & 64 & .0005 & Norm 
 & & 8  & 3--\phantom{{0}}64 & Soft & 64 & .0010 & Unif 
 & & 30  & 3--512 & Relu & 32 & .0005 & Lapl 
 \\ 
Wine 
 & & 2--\phantom{{0}}64 & Relu & 16 & .0010 & Norm 
 & & 8  & 3--\phantom{{0}}64 & Relu & 16 & .0010 & Norm 
 & & 30  & 3--\phantom{{0}}64 & Relu & 16 & .0010 & Lapl 
 \\ 
Concrete 
 & & 3--512 & Relu & 64 & .0005 & Unif 
 & & 12  & 3--512 & Relu & 16 & .0010 & Norm 
 & & 20  & 4--512 & Relu & 16 & .0010 & Norm 
 \\ 
Energy 
 & & 3--512 & Relu & 16 & .0010 & Norm 
 & & 8  & 3--512 & Relu & 16 & .0005 & Unif 
 & & 30  & 3--512 & Soft & 64 & .0005 & Unif 
 \\ 
Boston 
 & & 3--512 & Relu & 16 & .0010 & Norm 
 & & 12  & 3--512 & Relu & 32 & .0010 & Norm 
 & & 30  & 2--512 & Soft & 256 & .0010 & Unif 
 \\ 
Yacht 
 & & 4--512 & Relu & 16 & .0005 & Unif 
 & & 8  & 2--512 & Relu & 64 & .0010 & Unif 
 & & 30  & 4--512 & Relu & 16 & .0010 & Norm 
 \\ 
\bottomrule 
\end{tabular} 